\documentclass{article}

\usepackage{silence}
\PassOptionsToPackage{numbers, compress}{natbib}

\usepackage[final]{staix_2026_arxiv}

\usepackage[utf8]{inputenc}
\usepackage[T1]{fontenc}
\usepackage{hyperref}
\usepackage{url}
\usepackage{booktabs}
\usepackage{amsfonts}
\usepackage{nicefrac}
\usepackage{microtype}
\usepackage{xcolor}
\usepackage{graphicx}
\usepackage{amsmath,amssymb,amsthm}

\usepackage{mathtools}
\usepackage{bm}
\usepackage{enumitem}
\usepackage{subcaption}
\usepackage{tikz}
\usepackage{wrapfig}
\usepackage{array}
\usepackage{multirow}
\usepackage{placeins}
\usepackage{float}
\usepackage{flafter}
\usepackage[nameinlink,capitalise,noabbrev]{cleveref}

\usepackage{crossreftools}
\usetikzlibrary{arrows.meta,positioning,fit,calc,shapes.geometric}

\newtheorem{theorem}{Theorem}
\newtheorem{proposition}[theorem]{Proposition}

\theoremstyle{definition}

\theoremstyle{remark}

\crefname{theorem}{Theorem}{Theorems}
\Crefname{theorem}{Theorem}{Theorems}
\crefname{proposition}{Proposition}{Propositions}
\Crefname{proposition}{Proposition}{Propositions}
\crefname{lemma}{Lemma}{Lemmas}
\Crefname{lemma}{Lemma}{Lemmas}
\crefname{corollary}{Corollary}{Corollaries}
\Crefname{corollary}{Corollary}{Corollaries}
\crefname{assumption}{Assumption}{Assumptions}
\Crefname{assumption}{Assumption}{Assumptions}

\newcommand{\R}{\mathbb{R}}
\newcommand{\E}{\mathbb{E}}

\newcommand{\N}{\mathcal{N}}
\newcommand{\cD}{\mathcal{D}}
\newcommand{\cL}{\mathcal{L}}
\newcommand{\cR}{\mathcal{R}}
\newcommand{\cE}{\mathcal{E}}
\newcommand{\cW}{\mathcal{W}}
\newcommand{\KL}{\mathrm{KL}}
\newcommand{\Law}{\mathrm{Law}}
\newcommand{\tr}{\mathrm{tr}}
\newcommand{\diag}{\mathrm{diag}}

\newcommand{\pd}{\mathrm{pd}}
\newcommand{\sd}{\mathrm{sd}}

\newcommand{\RF}{\mathrm{RF}}

\newcommand{\F}{\mathrm{F}}
\newcommand{\norm}[1]{\left\lVert #1\right\rVert}
\newcommand{\inner}[2]{\left\langle #1,#2\right\rangle}
\newcommand{\dd}{\mathrm{d}}

\newcommand{\Umat}{U}
\newcommand{\Uhat}{\widehat U}
\newcommand{\eps}{\varepsilon}

\usepackage{mathrsfs}

\def\cD{\mathcal{D}}
\def\cE{\mathcal{E}}

\def\cL{\mathcal{L}}

\def\cR{\mathcal{R}}

\def\cW{\mathcal{W}}

\def\cZ{\mathcal{Z}}

\usepackage{yfonts}

\def\KL{{\sf KL}}

\def\tr{{\sf Tr}}

\usepackage{bbm}

\mathchardef\mhyphen="2D

\def\Rone{{\rm I}}
\def\Rtwo{{\rm II}}
\def\Rthree{{\rm III}}
\usepackage{algorithm}
\usepackage{algorithmicx}
\RequirePackage{algpseudocode}

\usepackage{tocbibind}
\usepackage{microtype}
\usepackage{titletoc}

\usepackage[textsize=tiny]{todonotes}

\usepackage[dvipsnames]{xcolor}%
\usepackage{hyperref}       %
\hypersetup{
    colorlinks=true,
    linkcolor=BrickRed,
    filecolor=magenta,      
    urlcolor=magenta,
    citecolor= OliveGreen
}

\newcommand{\Wtwo}{\mathcal W_2}

\newcommand{\argmin}{\mathop{\mathrm{argmin}}}

\title{Optimal Self-Distillation for Rectified Flow\\via Linear Probing}

\author{%
  Saptarshi Roy \\
  University of Texas, Austin \\
  \texttt{saptarshiroy@utexas.edu}
  \And
  Debepsita Mukherjee \\
  University of Texas, Austin \\
  \texttt{debepsitamukherjee@utexas.edu}
  \And
  Pratik Patil \\
  University of Texas, Austin \\
  \texttt{pratikpatil@utexas.edu}
}

\begin{document}

\maketitle

\begin{abstract}
Modern generative models are increasingly trained using model-generated signals, creating both opportunities for self-improvement and risks of collapse.
We study optimal self-distillation (SD) for rectified flow (RF): given a suboptimal teacher velocity field, can a student trained on a mixture of true RF velocities and teacher velocities provably improve the teacher?
For linear RF with ridge regularization on fixed interpolation pairs, we prove an exact affine path identity, derive the optimal mixing coefficient in closed form, and show strict improvement in integrated velocity risk whenever the teacher risk is nonstationary along the regularization path.
The optimal coefficient obeys a sign rule: positive mixing corrects under-regularized teachers, while negative mixing corrects over-regularized teachers.
We also give one-shot generalized cross-validation (GCV) and validation tuning procedure that avoids grid search over mixing weights and repeated refitting.
Combining this theorem with RF Wasserstein convergence bounds, we show that optimal self-distillation improves the velocity estimation terms controlling continuous-time and finite-step generation error.
Experiments with Gaussian models, Gaussian mixtures, and image data show that optimal self-distillation improves velocity risk, mode recovery, and finite-step generation relative to both the teacher and pure distillation.
\end{abstract}

\begin{figure*}[t]
    \centering
    \includegraphics[width=0.8\textwidth]{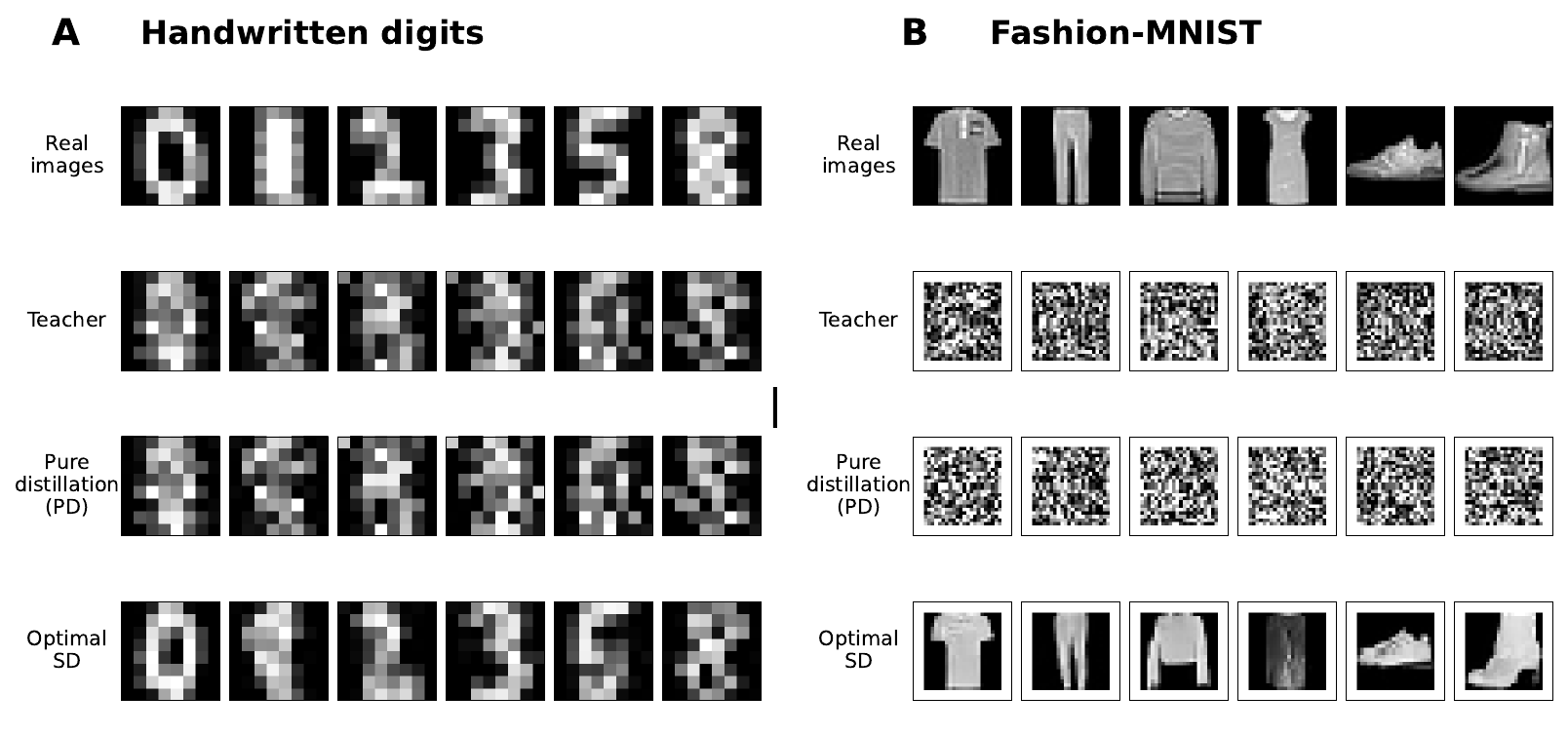}
    \caption{
    \textbf{Optimal self-distillation corrects degraded RF teachers.}
    In controlled stress tests on real images from handwritten digits and Fashion-MNIST, we scale down the teacher output to emulate an over-regularized or otherwise miscalibrated teacher field.
    The resulting teacher and pure-distilled student produce degraded samples, whereas validation selects a negative mixing coefficient for unconstrained SD, which recovers recognizable samples.
    The corresponding validation RF risks decrease from $1.237$ to $0.415$ on digits and from $0.828$ to $0.210$ on Fashion-MNIST.
    }
    \label{fig:intro-selling}
    \vspace{-1em}
\end{figure*}

\section{Introduction}
\label{sec:introduction}

Synthetic data and model-generated supervision are increasingly common in modern AI pipelines, with seemingly opposing consequences.
Recursively training generative models on their own outputs can lead to model collapse, loss of diversity, or distributional drift \citep{shumailov2024ai,fu2024towards,zhu2024analyzing}.
Yet model-generated signals can also be useful: in supervised learning, self-distillation often improves a model by retraining on a mixture of ground-truth labels and teacher predictions.
We ask whether this positive self-improvement phenomenon can be isolated and analyzed in a generative model.

\enlargethispage{\baselineskip}

We focus on rectified flow (RF) within the broader flow matching framework \citep{lipman2023flow,liu2023flow,albergo2023building}, which offers a simple approach to generative modeling with strong empirical performance \citep{gen_esser2024scaling,gen_jin2025pyramidal,gen_le2023voicebox}.
Let $X_0\sim \rho_0$ be source noise and $X_1\sim \rho_1$ be data, and define
\begin{equation}
\label{eq:linear-interp}
    X_t :=(1-t)X_0+tX_1, \qquad u=X_1-X_0.
\end{equation}
RF learns a time-conditioned velocity field $v(t,x)$ by regressing $u$ on $(t,X_t)$ with squared loss, thereby approximating a suitable velocity field $v^\star(t,x)$.
This problem has the same geometry as supervised regression, but the learned field subsequently defines an ODE sampler.
RF is therefore a natural setting in which to ask whether a student trained on mixed teacher-generated and true velocity targets can provably improve a suboptimal RF teacher.

The answer is yes for a linear RF model trained on fixed interpolants. We adopt \textit{linear probing} \citep{kumar2022finetuning_probing,tomihari2024understanding_probing}, an efficient adaptation strategy that freezes a pretrained representation or feature map and updates only a lightweight linear head $W$.
We keep the interpolation covariates fixed, fit a teacher $\widehat W_\lambda$ with ridge regularization on the true velocity targets, evaluate it at those interpolation points, and train a student on mixed targets (see Figure~\ref{fig:schematic}).
For a true velocity $u$, a teacher prediction $\widehat u$, and a mixing coefficient $\xi\in\R$, the mixed target is $u^{(\xi)}=(1-\xi)u+\xi\widehat u$.
Unlike recursive Reflow, which generates new noise--data pairs and changes the endpoint coupling, keeping the interpolants fixed preserves the ridge geometry: the self-distilled student lies on an affine path between the teacher and a pure-distilled refit.
Consequently, the integrated RF risk is an exact quadratic in $\xi$, yielding an optimum in closed form and a theorem guaranteeing strict improvement.

\subsection{Contributions and outline}

\begin{figure}[!t]
\centering
\begin{tikzpicture}[font=\small, node distance=0.45cm and 0.7cm, >=Latex]
\tikzstyle{box}=[draw, rounded corners=2pt, align=center, minimum height=0.8cm, inner sep=4pt, fill=gray!6]
\tikzstyle{bluebox}=[box, fill=blue!6, draw=blue!50]
\tikzstyle{greenbox}=[box, fill=green!6, draw=green!45!black]
\tikzstyle{orangebox}=[box, fill=orange!9, draw=orange!70!black]
\node[box] (data) {Fixed RF pairs\\$(T_i,X_{T_i,i})$};
\node[bluebox, right=of data] (true) {True velocity\\$u_i=X_{1,i}-X_{0,i}$};
\node[orangebox, below=of true] (teach) {Teacher velocity\\$\widehat u_i=v_{\widehat W_\lambda}(T_i,X_{T_i,i})$};
\node[greenbox, right=1.1cm of true, yshift=-0.45cm] (mix) {Mixed velocity target\\$u_i^{(\xi)}=(1-\xi)u_i+\xi\widehat u_i$};
\node[greenbox, right=of mix] (student) {Student\\$\widehat W_{\sd,\lambda,\xi}$};
\node[box, below=of student] (risk) {Choose $\xi$ to minimize\\integrated RF risk};
\draw[->] (data) -- (true);
\draw[->] (data) |- (teach);
\draw[->] (true) -- (mix);
\draw[->] (teach) -- (mix);
\draw[->] (mix) -- (student);
\draw[->] (student) -- (risk);
\node[align=center, below=1cm of data, text width=3.0cm] {Fixed interpolants};
\end{tikzpicture}
\caption{\textbf{Schematic of self-distillation for RF velocity targets on fixed interpolants.} Unlike recursive Reflow, the endpoint coupling and interpolation covariates remain fixed. The teacher predicts velocities at the same RF training locations, and the student is trained on their mixture with the true targets.}
\label{fig:schematic}
\vspace{-1em}
\end{figure}
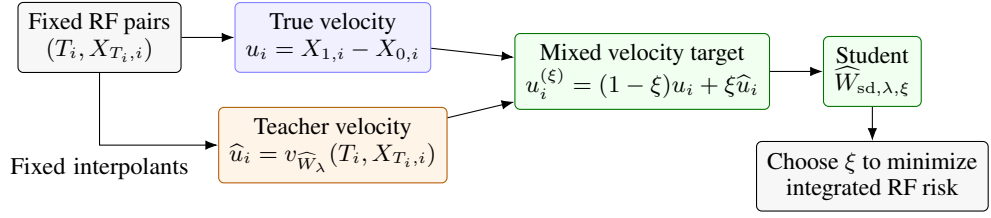

\begin{enumerate}[leftmargin=2em]

\item
\textbf{Optimal self-distillation for RF velocity learning.}
Section~\ref{sec:sd-velocity-learning} introduces self-distillation of RF velocity targets on fixed interpolants; Figure~\ref{fig:schematic} and Algorithm~\ref{alg:rf-self-distillation} summarize the procedure.
For linear RF with ridge regularization, we prove an exact affine path identity linking the teacher, pure-distilled refit, and self-distilled student (Proposition~\ref{prop:affine}).
This identity makes the integrated RF risk an exact quadratic in $\xi$, yielding an optimum in closed form and strict improvement whenever the teacher risk is nonstationary along the regularization path (Theorem~\ref{thm:optimal-sd}).
We also propose one-shot tuning procedure based on GCV or validation that avoids grid search and repeated refitting, providing an efficient adaptation strategy for RF models based on fixed features or linear probing.

\item
\textbf{Consequences for generation error.}
Section~\ref{sec:generative-consequences} connects the improvement in velocity risk to generative performance.
RF regression risk decomposes into excess velocity approximation error and an irreducible term, so the strict reduction in RF risk also reduces the velocity error quantity entering RF sampling bounds.
Combining this identity with Wasserstein convergence results for RF, we show that optimal SD improves the continuous-time generation error upper bound (Proposition~\ref{prop:continuous-generation-bound}) and the velocity estimation term in finite-step Euler bounds (Theorem~\ref{thm:w2-ode-disc}).

\item
\textbf{Empirical validation.}
Section~\ref{sec:experiments} validates the theory using Gaussian models, nonlinear Gaussian mixtures, random features, and controlled RF experiments on real images.
Across these settings, optimal SD improves velocity risk and finite-step generation, one-shot tuning closely tracks the oracle choice, and unconstrained negative mixing is essential for correcting over-regularized teachers; see Figures~\ref{fig:intro-selling}, \ref{fig:gaussian-main}, and \ref{fig:mog-scatter}, as well as Table~\ref{tab:cifar-neural-rebuttal}.
A time-conditioned U-Net experiment on CIFAR-10 tests the same correction beyond linear probing with fixed features by comparing the teacher, pure distillation, clipped SD, and unconstrained SD.
The supplement reports metric definitions, sensitivity to teacher quality and output scaling, and a fine-tuning baseline with matched compute.
\end{enumerate}

\subsection{Related work}
\label{sec:related}

\textbf{Rectified flow and flow matching.}
Flow matching and stochastic interpolants learn velocity fields by squared regression against conditional velocity targets \citep{albergo2023building,lipman2023flow}.
Couplings inspired by optimal transport and careful path design can produce straighter flows, stabilize training, and accelerate sampling \citep{pooladian2023multisample,shaul2023kinetic,tong2023improving}; complementary deterministic sampling error bounds depend on velocity field approximation error and flow regularity \citep{benton2023error}.
RF specializes this framework to linear interpolants and uses the learned ODE to transport source noise to the data distribution \citep{liu2023flow}.
A Wasserstein analysis of RF further shows that terminal sampling error is controlled by integrated velocity estimation error and an Euler discretization term governed by straightness \citep{bansal2026on}.
We use this regression view of estimation error to analyze optimal self-distillation of the learned velocity field.

\textbf{Self-consuming generative models and collapse.}
Model collapse under recursive synthetic data training has been observed and studied across a range of settings \citep{hataya2023will,martinez2023combining,bohacek2023nepotistically,shumailov2024ai,fu2024towards,alemohammad2024self}.
In RF, recursive Reflow repeatedly changes the endpoint coupling; proposed safeguards include mixing real and synthetic endpoint pairs \citep{zhu2024analyzing} and \textit{conic reflow}, which learns a spherical linear interpolation between real data and its inverse noise \citep{kwon2026balanced}.
Our complementary setting keeps the endpoint coupling and interpolation covariates fixed, isolating a one-shot target mixing mechanism that is analytically tractable and computationally efficient.

\textbf{Self-distillation and optimal mixing.}
Self-distillation retrains a student on the same examples using a mixture of ground-truth labels and teacher predictions.
Classical distillation uses soft teacher predictions to train a student, often transferring teacher performance to a smaller or retrained model \citep{hinton2015distilling,furlanello2018born,phuong2019towards,zhang2019your,ji2020knowledge}.
Theory explains self-distillation gains through implicit regularization, bias--variance tradeoffs, teacher mimicry, and repeated distillation effects \citep{mobahi2020self,das2023understanding,pareek2024understanding}.
Related work also studies uncertainty-aware Bayesian knowledge distillation \citep{fang2024bayesian}; \citet{luo2023diffusiondistill} survey distillation methods for diffusion models.
For ridge regression, \citet{dang2026optimal} show that unconstrained optimal mixing can strictly improve any nonstationary teacher along the ridge path and that the optimal coefficient can be negative in over-regularized regimes.
We transfer this structural geometry from supervised ridge regression to RF velocity matching.

\section{Rectified flow}
\label{sec:setup}

Let $\rho_0$ be the source distribution, typically $\N(0,I_d)$, and $\rho_1$ the data distribution.
We sample endpoint pairs $(X_0,X_1)$ from a coupling $\pi$ of $(\rho_0,\rho_1)$, usually the independent coupling $\pi=\rho_0\otimes\rho_1$.
For $t\in[0,1]$, define the linear interpolant and pairwise velocity as in \eqref{eq:linear-interp}.
RF learns a time-dependent velocity field by minimizing
\begin{equation}
    \label{eq:rf-loss}
    \cL_{\RF}(v)
    =\E_{T\sim\nu,\,(X_0,X_1)\sim\pi}
    \big[\norm{u-v(T,X_T)}^2\big],
\end{equation}
where $\nu$ is usually uniform on $[0,1]$.
The population minimizer is $v^\star(t,x)=\E[u\mid X_t=x]$.
If $Z_t$ solves
\begin{equation}
\label{eq:true-rf-ode}
    \frac{\dd Z_t}{\dd t}=v^\star(t,Z_t),\qquad Z_0\sim\rho_0,
\end{equation}
then $Z_t$ has the same marginal law as $X_t$ for every $t\in[0,1]$ \citep{liu2023flow}.
The projection identity under squared loss gives
\begin{equation}
    \label{eq:irreducible}
    \E\norm{u-v(T,X_T)}^2
    =
    \E\norm{v^\star(T,X_T)-v(T,X_T)}^2
    +
    \E\norm{u-v^\star(T,X_T)}^2.
\end{equation}
Thus RF regression risk equals integrated velocity approximation error plus an irreducible conditional variance term.

We study the linear (in parameters) model
\begin{equation}
    \label{eq:linear-model}
    v_W(t,x)=W^\top\phi(t,x),\qquad W\in\R^{m\times d},
\end{equation}
where $\phi:[0,1]\times\R^d\to\R^m$ is a fixed time-conditioned feature map.
For example, let $b(t)=(b_1(t),\ldots,b_L(t))^\top$ be fixed temporal basis functions and set $\phi(t,x)=b(t)\otimes x$.
Writing $W=(A_1^\top,\ldots,A_L^\top)^\top$ with $A_\ell\in\R^{d\times d}$ gives $v_W(t,x)=A_W(t)x$, where $A_W(t):=\sum_{\ell=1}^L b_\ell(t)A_\ell$.
This yields a time-dependent linear velocity field with parameters shared across time.
More generally, the fixed map $\phi$ may comprise random features, kernel or Nystr\"om approximations, or frozen pretrained representations \citep{du2024implicit}; the affine path identity below applies to any fixed choice of $\phi$.
Once $\phi$ is fixed, training the RF teacher reduces to estimating the shared coefficient matrix $W$.

Given training triples $\cD_N=\{(X_{0,i},X_{1,i},T_i)\}_{i=1}^N$ with $T_i\stackrel{\mathrm{iid}}{\sim}\nu$, define $X_{T_i,i}=(1-T_i)X_{0,i}+T_iX_{1,i}$.
Let $u_i=X_{1,i}-X_{0,i}$ and $\phi_i=\phi(T_i,X_{T_i,i})$.
We use plain italic notation throughout: $u$ and $u_i$ are velocity vectors, while $\Umat$ denotes the stacked velocity target matrix.
Specifically, $\Phi=(\phi_1,\ldots,\phi_N)^\top\in\R^{N\times m}$ and $\Umat=(u_1,\ldots,u_N)^\top\in\R^{N\times d}$.
The RF teacher fitted with ridge regularization is
\begin{equation}
    \label{eq:teacher}
    \widehat W_\lambda
    =\argmin_W \, \frac1N\norm{\Umat-\Phi W}_\F^2+\lambda\norm{W}_\F^2
    =\Big(\frac{\Phi^\top\Phi}{N}+\lambda I_m\Big)^{-1}\frac{\Phi^\top\Umat}{N}.
\end{equation}
For any learned velocity field $v$, define the conditional integrated RF risk
\begin{equation}
    \label{eq:conditional-risk}
    \cR_\nu(v\mid\cD_N)
    =\E\big[\norm{u-v(T,X_T)}^2\mid \cD_N\big],
\end{equation}
where the expectation is over an independent test triple with $T\sim\nu$ and $(X_0,X_1)\sim\pi$.
For a linear estimator $\widehat W$, we use the shorthand $\cR_\nu(\widehat W\mid\cD_N):=\cR_\nu(v_{\widehat W}\mid\cD_N)$.
When $\nu$ is uniform on $[0,1]$, we omit the subscript and write $\cR(v\mid\cD_N)$.

\section{Optimal self-distillation}
\label{sec:sd-velocity-learning}

The teacher's fitted velocities at the original RF interpolation points form $\Uhat_\lambda=\Phi\widehat W_\lambda$.
Pure distillation retrains on these targets:
\begin{equation}
    \label{eq:pd}
    \widehat W_{\pd,\lambda}
    =\argmin_{W} \, \frac1N\norm{\Uhat_\lambda-\Phi W}_\F^2+\lambda\norm{W}_\F^2.
\end{equation}
For $\xi\in\R$, define the mixed velocity target matrix
\begin{equation}
    \Umat^{(\xi)}=(1-\xi)\Umat+\xi\Uhat_\lambda,
    \label{eq:mixed-target}
\end{equation}
and let $\widehat W_{\sd,\lambda,\xi}$ be the ridge RF fit on $(\Phi,\Umat^{(\xi)})$.
For compactness, throughout the linear theory we write
\[
    v_\lambda:=v_{\widehat W_\lambda},\qquad
    v_{\pd,\lambda}:=v_{\widehat W_{\pd,\lambda}},\qquad
    v_{\sd,\lambda,\xi}:=v_{\widehat W_{\sd,\lambda,\xi}}.
\]

\begin{figure}[!t]
\centering
\begin{subfigure}[t]{0.49\linewidth}
\centering
\includegraphics[width=\linewidth]{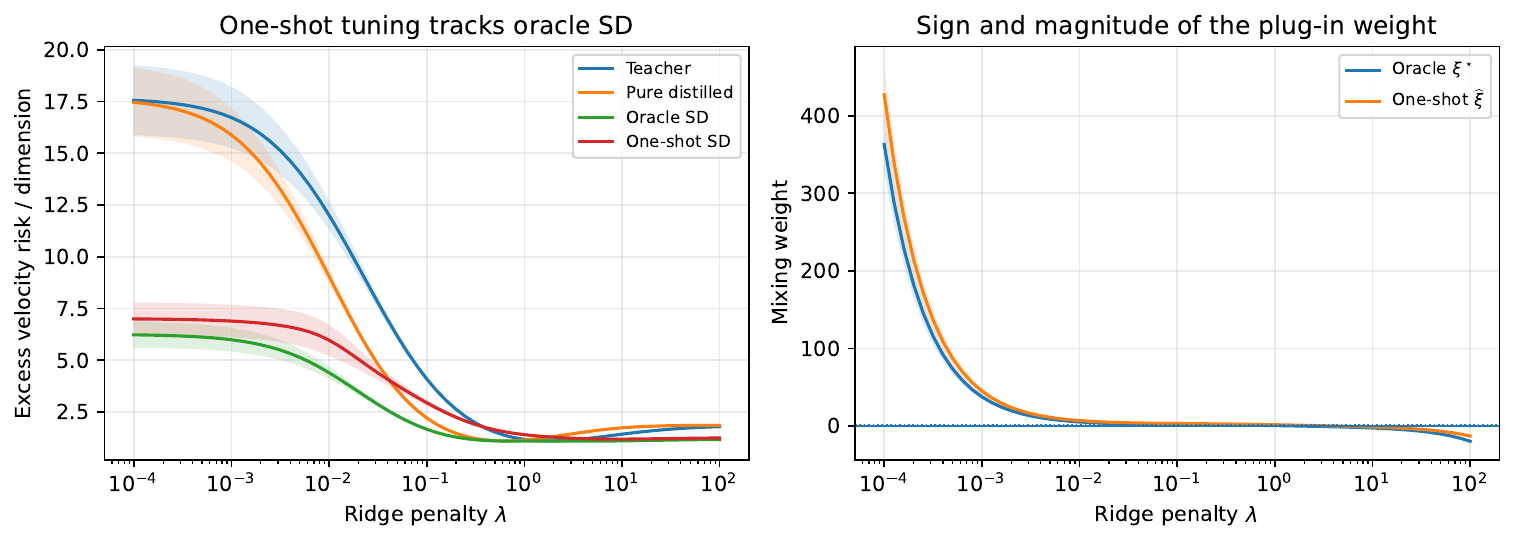}
\caption{One-shot GCV tuning.}
\label{fig:gcv}
\end{subfigure}\hfill
\begin{subfigure}[t]{0.49\linewidth}
\centering
\includegraphics[width=\linewidth]{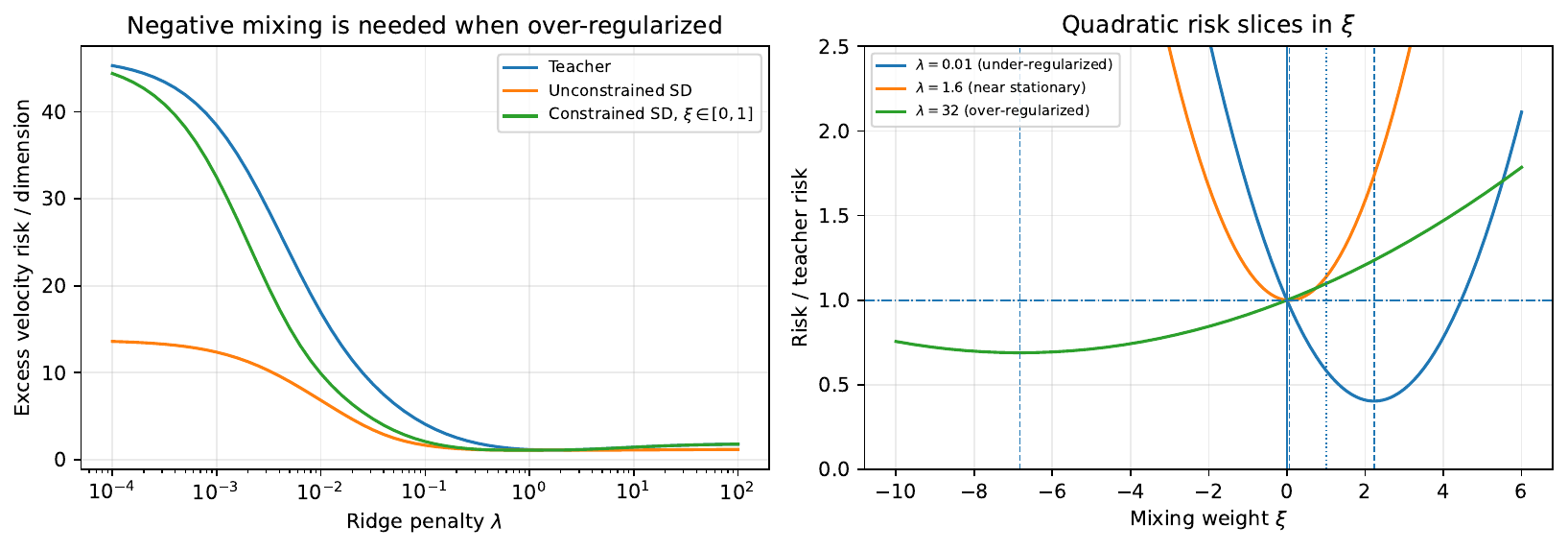}
\caption{Unconstrained mixing is necessary.}
\label{fig:constrained}
\end{subfigure}
\caption{\textbf{The sign rule in Gaussian RF.} (a) In a correctly specified Gaussian RF model, one-shot GCV closely tracks oracle mixing and selects negative mixing in over-regularized regimes. (b) Constraining $\xi\in[0,1]$ often returns the teacher when the optimal correction is negative.}
\label{fig:gaussian-main}
\vspace{-1em}
\end{figure}

\subsection{Oracle optimal mixing}
\label{subsec:oracle-mixing}

\begin{proposition}[Affine path]
\label{prop:affine}
For every $\lambda>0$ and $\xi\in\R$,
\begin{equation}
    \widehat W_{\sd,\lambda,\xi}
    =
    (1-\xi)\widehat W_\lambda+\xi\widehat W_{\pd,\lambda}.
    \label{eq:affine-path}
\end{equation}
Consequently, for every $(t,x)$,
\[
    v_{\sd,\lambda,\xi}(t,x)
    =
    (1-\xi)v_\lambda(t,x)+\xi v_{\pd,\lambda}(t,x).
\]
\end{proposition}

The affine family in Proposition~\ref{prop:affine} is reminiscent of classifier-free guidance (CFG) in diffusion and flow models \citep{cfg_ho2021classifier,cfg_saini2025rectified}, but the endpoints and purpose differ. CFG typically extrapolates from an unconditional field toward a conditional field to strengthen conditioning. Here the path joins a teacher trained on true targets and its pure-distilled refit; when $\xi<0$, the estimator moves away from the pure-distilled field to correct repeated shrinkage. Negative guidance has also been explored in diffusion models for suppressing unwanted features \citep{koulischer2025dynamic}.
For notational simplicity, define
\begin{equation}
\label{eq:r-definitions}
    R(\lambda)=\cR(\widehat W_\lambda\mid\cD_N),\quad
    R_{\pd}(\lambda)=\cR(\widehat W_{\pd,\lambda}\mid\cD_N),\quad
    R_{\sd}(\lambda,\xi)=\cR(\widehat W_{\sd,\lambda,\xi}\mid\cD_N).
\end{equation}
Let $(T,X_T,u)$ denote an independent test triple and define
\begin{align}
    C(\lambda)
    &=
    \E\big[
        \inner{u-v_\lambda(T,X_T)}
        {u-v_{\pd,\lambda}(T,X_T)}
        \mid\cD_N
    \big],
    \label{eq:C}\\
    D(\lambda)
    &=
    \E\big[
        \norm{v_\lambda(T,X_T)-v_{\pd,\lambda}(T,X_T)}^2
        \mid\cD_N
    \big].
    \label{eq:D}
\end{align}
By expansion, $D(\lambda)=R(\lambda)+R_{\pd}(\lambda)-2C(\lambda)\ge 0$.

\begin{theorem}[Optimal self-distillation for integrated RF risk]
\label{thm:optimal-sd}
Fix $\lambda>0$ and assume $D(\lambda)>0$. Then, for every $\xi\in\R$,
\begin{equation}
    R_{\sd}(\lambda,\xi)
    =
    R(\lambda)
    -
    2\xi\{R(\lambda)-C(\lambda)\}
    +
    \xi^2D(\lambda).
    \label{eq:risk-quadratic}
\end{equation}
Consequently,
\begin{equation}
    \xi^\star(\lambda)
    =
    \frac{R(\lambda)-C(\lambda)}{D(\lambda)},
    \qquad
    R_{\sd}^\star(\lambda)
    =
    R(\lambda)
    -
    \frac{(R(\lambda)-C(\lambda))^2}{D(\lambda)}.
    \label{eq:optimal-xi}
\end{equation}
Moreover, if $R(\lambda)$ is differentiable along the ridge path, then
\begin{equation}
    \xi^\star(\lambda)
    =
    -\frac{\lambda}{2}\frac{R'(\lambda)}{D(\lambda)},
    \qquad
    R_{\sd}^\star(\lambda)
    =
    R(\lambda)
    -
    \frac{\lambda^2}{4}\frac{(R'(\lambda))^2}{D(\lambda)}.
    \label{eq:sign-rule}
\end{equation}
Thus, whenever $R'(\lambda)\neq0$, $R_{\sd}^\star(\lambda)<R(\lambda)$ and
$\mathrm{sign}(\xi^\star(\lambda))=-\mathrm{sign}(R'(\lambda))$.
\end{theorem}

The sign rule determines the correction direction: on the increasing side of the ridge risk curve, $R'(\lambda)>0$ and $\xi^\star(\lambda)<0$, so the estimator extrapolates away from the pure-distilled field; on the decreasing side, $\xi^\star(\lambda)>0$.
Although $\xi<0$ makes \eqref{eq:mixed-target} an affine rather than convex target mixture, fitting the student remains an ordinary ridge regression problem for any real $\xi$.
Appendix~\ref{sec:time-varying-mixing} gives the time-dependent extension, its exact gain, and numerical comparisons.

\subsection{Data-dependent tuning}

For the linear ridge model, define the smoother matrix
\begin{equation}
    S_\lambda
    =
    \Phi
    \Big(\frac{\Phi^\top\Phi}{N}+\lambda I_m\Big)^{-1}
    \frac{\Phi^\top}{N}.
\end{equation}
Then $\Uhat_\lambda=S_\lambda\Umat$ and
$\Uhat_{\pd,\lambda}:=\Phi\widehat W_{\pd,\lambda}=S_\lambda^2\Umat$.
Following generalized cross-validation (GCV) for ridge regression (see, e.g., \citep{patil2021uniform,patil2022estimating}), we estimate $R$, $R_{\pd}$, and $C$ by
\begin{equation*}
    \widehat R
    =
    \frac{\norm{\Umat-S_\lambda\Umat}_\F^2/N}
    {\{1-\tr(S_\lambda)/N\}^2},
    \;
    \widehat R_{\pd}
    =
    \frac{\norm{\Umat-S_\lambda^2\Umat}_\F^2/N}
    {\{1-\tr(S_\lambda^2)/N\}^2},
    \;
    \widehat C
    =
    \frac{
    \inner{\Umat-S_\lambda\Umat}{\Umat-S_\lambda^2\Umat}_\F/N
    }
    {
    \{1-\tr(S_\lambda)/N\}
    \{1-\tr(S_\lambda^2)/N\}
    } ,
\end{equation*}
where $\inner{A}{B}_\F=\tr(A^\top B)$.
Plugging these estimates into \eqref{eq:optimal-xi} gives
\begin{equation}
    \widehat\xi_{\mathrm{GCV}}
    =
    \frac{\widehat R-\widehat C}
    {\widehat R+\widehat R_{\pd}-2\widehat C}.
    \label{eq:gcv-xi}
\end{equation}

\begin{algorithm}[!t]
\caption{One-shot RF self-distillation on fixed interpolants}
\label{alg:rf-self-distillation}
\begin{algorithmic}[1]
    \State Sample RF triples $(X_{0,i},X_{1,i},T_i)$ and form $X_{T_i,i}$ and $u_i=X_{1,i}-X_{0,i}$.
    \State Train a teacher \(v\) on true velocities \(u_i\).
    \State Form teacher velocities \(\widehat u_i=v(T_i,X_{T_i,i})\) and train a pure-distilled model \(v_{\pd}\) on \(\widehat u_i\).
    \State Estimate \(\xi\) by GCV or validation on the family \(v+\xi(v_{\pd}-v)\).
    \State Use \(v_{\widehat\xi}\), or train one student on \((1-\widehat\xi)u_i+\widehat\xi\,\widehat u_i\).
\end{algorithmic}
\end{algorithm}

\vspace{-1em}
For neural or other nonlinear RF models, train a teacher field $v$ and a pure-distilled field $v_{\pd}$ once, then choose $\xi$ by validation over the affine family
\begin{equation}
    v_\xi(t,x)
    =
    v(t,x)+\xi\{v_{\pd}(t,x)-v(t,x)\}.
    \label{eq:validation-mixture}
\end{equation}
This one-dimensional search avoids retraining a student for every candidate $\xi$.
Algorithm~\ref{alg:rf-self-distillation} summarizes the resulting one-shot procedure.

\section{Generative consequences}
\label{sec:generative-consequences}

This section shows how the improvement in velocity risk from optimal self-distillation tightens standard Wasserstein generation error bounds, both in continuous time and for Euler sampling.

\subsection{Generation error upper bounds}
\label{subsec:generation-bounds}

Let $v$ be any learned velocity field and let $\rho_v=\Law(Y_1^v)$, where
\begin{equation}
\label{eq:ode-y-v}
    \frac{\dd Y_t^v}{\dd t}=v(t,Y_t^v),\qquad Y_0^v\sim\rho_0 .
\end{equation}
For each time $t$, define the excess velocity approximation error
\begin{equation}
    \label{eq:excess-velocity-error}
    \cE_t(v)=\E\big[\norm{v^\star(t,X_t)-v(t,X_t)}^2\big],\qquad
    \cE(v)=\int_0^1 \cE_t(v)\,\dd t.
\end{equation}
By \eqref{eq:irreducible}, $\cR(v\mid\cD_N)$ and $\cE(v)$ differ only by the irreducible term $\E\norm{u-v^\star(T,X_T)}^2$, which is independent of $v$. Thus, any decrease in RF regression risk yields the same decrease in $\cE(v)$.

\begin{figure}[!t]
\centering
\includegraphics[width=0.99\linewidth]{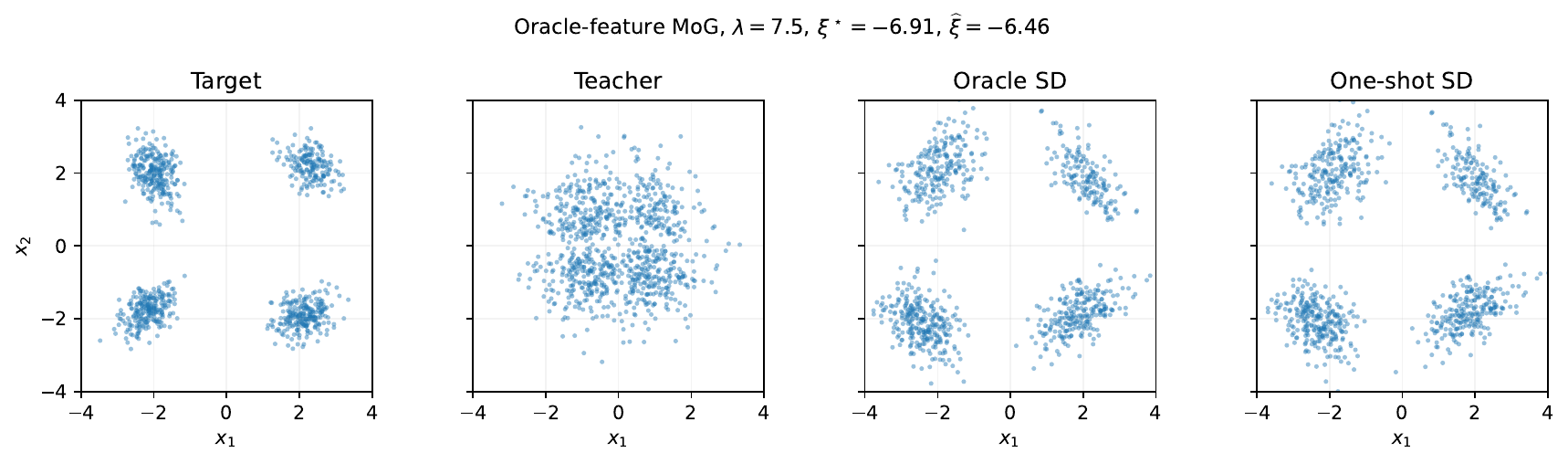}
\caption{
\textbf{Generated samples for nonlinear Gaussian mixture RF.}
At $\lambda\approx7.5$, the over-regularized teacher blurs the mixture.
Oracle and one-shot self-distillation use negative mixing and recover the target modes much more accurately.
}
\label{fig:mog-scatter}
\vspace{-1em}
\end{figure}

The next proposition combines this identity with a standard RF Wasserstein bound.
\begin{proposition}[Continuous-time generation bound improvement]
\label{prop:continuous-generation-bound}
Assume that $v(t,\cdot)$ is $L$-Lipschitz in $x$, uniformly over $t$, and that the ODEs \eqref{eq:true-rf-ode} and \eqref{eq:ode-y-v} have unique solutions. Then
\begin{equation}
    \cW_2^2(\rho_v,\rho_1)
    \le \exp(1+2L)\,\cE(v).
    \label{eq:continuous-w2-bound}
\end{equation}
Consequently, for $v_{\sd}^\star:=v_{\sd,\lambda,\xi^\star(\lambda)}$, the teacher $v_\lambda$, and the pure-distilled field $v_{\pd,\lambda}$,
\begin{equation}
    \cE(v_{\sd}^\star)\le \min\{\cE(v_\lambda),\cE(v_{\pd,\lambda})\},
\end{equation}
    and, using a common Lipschitz bound $L$ for these fields, the reduction relative to the teacher's bound is
\begin{equation}
    \exp(1+2L)\{\cE(v_\lambda)-\cE(v_{\sd}^\star)\}
    =\exp(1+2L)\frac{\lambda^2}{4}\frac{(R'(\lambda))^2}{D(\lambda)}.
    \label{eq:bound-gain-cont}
\end{equation}
\end{proposition}
The inequality in \eqref{eq:continuous-w2-bound} follows from \citet{bansal2026on}. Proposition~\ref{prop:continuous-generation-bound} combines it with Theorem~\ref{thm:optimal-sd}: optimal self-distillation weakly improves the continuous-time Wasserstein upper bound relative to both the teacher and pure distillation, with strict improvement over the teacher whenever $R'(\lambda)\ne0$.

\subsection{Guarantee for the discretized sampler}
For finite-step RF sampling, the same idea applies to the gridwise velocity errors.
Let $t_i=i/K$, $i=0,\ldots,K$, be the uniform Euler grid, and let $\widehat\rho_{v,K}$ be the law of the output of
\begin{equation}
\label{eq:emp-ode-disc}
    \widehat Y_{t_{i+1}}=\widehat Y_{t_i}+(t_{i+1}-t_i)v(t_i,\widehat Y_{t_i}),
    \qquad \widehat Y_0\sim\rho_0.
\end{equation}
At a fixed grid time $t_i$, the self-distilled conditional risk has the decomposition
\begin{equation}
\label{eq:r-ti}
    R_{\sd,t_i}(\lambda,\xi)=R_{t_i}(\lambda)-2\xi A_{t_i}(\lambda)+\xi^2 D_{t_i}(\lambda),
    \qquad A_{t_i}(\lambda):= R_{t_i}(\lambda)-C_{t_i}(\lambda),
\end{equation}
where $R_{\sd,t_i}(\lambda,\xi)$, $R_{t_i}(\lambda)$, $C_{t_i}(\lambda)$, and $D_{t_i}(\lambda)$ are defined as in \eqref{eq:r-definitions}, \eqref{eq:C}, \eqref{eq:D}, and \eqref{eq:risk-quadratic}, with expectations conditioned on $T=t_i$. Define the gridwise errors
\begin{equation}
\label{eq:error-definitions}
    \eps_i^2(v):=\cE_{t_i}(v),\qquad
    \bar\eps_K^2(v):=\frac1K\sum_{i=0}^{K-1}\eps_i^2(v).
\end{equation}
All sums below run over $i=0,\ldots,K-1$. When $\sum_iD_{t_i}(\lambda)>0$, the coefficient minimizing the average gridwise error and its gain are
\begin{equation}
\begin{aligned}
    \xi_{\mathrm U,K}^\star(\lambda)
    &=\frac{\sum_i A_{t_i}(\lambda)}{\sum_iD_{t_i}(\lambda)},\\
    \bar\eps_K^2(v_\lambda)-\bar\eps_K^2(v_{\sd,\lambda,\xi_{\mathrm U,K}^\star})
    &=\frac{\{\sum_i A_{t_i}(\lambda)\}^2}{K\sum_i D_{t_i}(\lambda)}.
\end{aligned}
\label{eq:grid-average-gain}
\end{equation}
Write $v_{\sd,K}^\star:=v_{\sd,\lambda,\xi_{\mathrm U,K}^\star(\lambda)}$.

\begin{figure}[!t]
\centering
\includegraphics[width=0.9\linewidth]{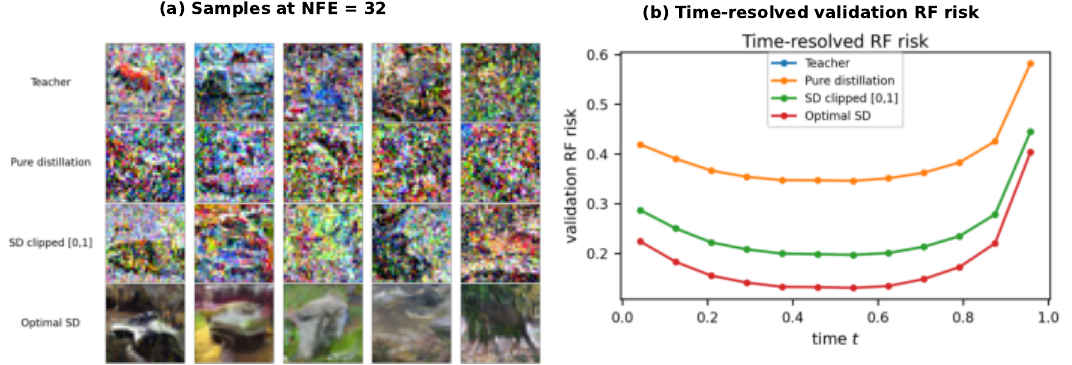}
\caption{\textbf{Neural CIFAR-10 RF experiment.} Left: the teacher and pure-distilled samples are degraded. Clipping $\widehat\xi$ to $[0,1]$ returns the teacher because validation selects $\widehat\xi<0$, whereas unconstrained optimal SD produces clearer samples. Right: optimal SD reduces RF risk across the sampled time grid.}
\vspace{-1em}
\label{fig:cifar-neural-main}
\end{figure}

Let $\cZ^\star:=\{Z_t\}_{t\in[0,1]}$ denote the exact RF trajectory generated by \eqref{eq:true-rf-ode}. Following \citet{bansal2026on}, define its piecewise straightness by
\[
\gamma_{2,K}(\cZ^\star)
:=\max_{0\le i<K}\frac{1}{t_{i+1}-t_i}
\int_{t_i}^{t_{i+1}}\E\norm{\frac{\dd}{\dd t}v^\star(t,Z_t)}_2^2\,\dd t.
\]
\begin{theorem}[Euler generation bound improvement]
    \label{thm:w2-ode-disc}
    Under the assumptions of Proposition~\ref{prop:continuous-generation-bound}, the sampling distribution $\widehat\rho_{v,K}$ produced by \eqref{eq:emp-ode-disc} satisfies, conditionally on $\cD_N$,
    \begin{equation}
    \label{eq:euler-w2-bound}
    \cW_2^2(\widehat\rho_{v,K},\rho_1) \le C_L\left\{\gamma_{2,K}(\cZ^\star)/K^2 + \bar\eps_K^2(v)\right\},
    \end{equation}
    where $C_L=6\exp(1+\sqrt{1+12L^2})/(1+\sqrt{1+12L^2})$.
If $\sum_iD_{t_i}(\lambda)>0$, then for the gridwise optimally mixed field $v_{\sd,K}^\star$, the teacher $v_\lambda$, and the pure-distilled field $v_{\pd,\lambda}$,
\begin{equation}
    \bar\eps_K^2(v_{\sd,K}^\star)
    \le \min\{\bar\eps_K^2(v_\lambda),\bar\eps_K^2(v_{\pd,\lambda})\},
\end{equation}
and, using a common Lipschitz bound $L$, the reduction in the velocity error term for the teacher is
\begin{equation}
    C_L\{\bar\eps_K^2(v_\lambda)-\bar\eps_K^2(v_{\sd,K}^\star)\}
    =C_L\frac{\{\sum_i A_{t_i}(\lambda)\}^2}{K\sum_i D_{t_i}(\lambda)}.
    \label{eq:emp-bound-gain-cont}
\end{equation}
\end{theorem}
Equation~\eqref{eq:grid-average-gain} is the gridwise analogue of \eqref{eq:bound-gain-cont}; whenever $\sum_i A_{t_i}(\lambda)\ne0$, it strictly reduces the velocity error term and hence the corresponding upper bound relative to the teacher.
Appendix~\ref{sec:proof-generation-error-discrete} proves the result; Appendix~\ref{sec:optimal-mixing-discrete-sampler} gives the gridwise derivation.
The straightness term in \eqref{eq:euler-w2-bound} is common to all learned estimators for the same target RF path \citep{bansal2026on}; self-distillation affects only the velocity approximation error on the grid.
These results establish an improvement in the RF velocity error component of standard generation error bounds.
To complement these general upper bounds, Appendix~\ref{app:gaussian} gives exact Wasserstein and KL errors in a Gaussian linear model and shows explicitly how they depend on the integrated RF excess risk.
Neither the bounds nor this Gaussian calculation imply that every downstream sample quality metric must strictly improve in every misspecified model: finite-step generation also depends on Lipschitz constants, straightness, ODE stability, discretization, and the expressivity of the learned field.
Accordingly, our experiments report both velocity risks and generation metrics.

\begin{table}[!t]
\centering
\caption{\textbf{Neural CIFAR-10 RF experiment.}
We scale a trained time-conditioned U-Net RF field by $a=0.75$ to create a teacher with a miscalibrated velocity scale. Pure distillation compounds the shrinkage, clipped SD returns the teacher, and unconstrained SD selects $\widehat\xi=-1.225$. Metrics use $10{,}000$ generated samples with $32$ function evaluations.}
\label{tab:cifar-neural-rebuttal}
\begin{tabular}{cccc}
\toprule
Method & RF risk $\downarrow$ & FID $\downarrow$ & KID $\downarrow$ \\
\midrule
Teacher & 0.2435 & 284.26 & 0.3215 \\
Pure distillation & 0.3885 & 324.77 & 0.3569 \\
SD clipped $[0,1]$ & 0.2435 & 283.56 & 0.3196 \\
Optimal SD & 0.1803 & 30.08 & 0.0186 \\
\bottomrule
\end{tabular}
\end{table}

\section{Numerical evaluations}
\label{sec:experiments}

We evaluate the theory in Gaussian and nonlinear mixture models, then examine finite-step generation and RF experiments on real images.
All experiments use fixed interpolants: the teacher and pure-distilled model are trained once, then $\xi$ is selected by oracle test risk (as a benchmark), GCV, or validation.
We report validation and test RF risk because it is the quantity controlled by theory and used in RF generation bounds.

\paragraph{Gaussian RF.}
In the correctly specified Gaussian setting $X_1\sim\N(0,4I_d)$, the population velocity is $v^\star(t,x)=g(t)x$ (see Appendix~\ref{app:gaussian}) and the feature map is exact. Figure~\ref{fig:gaussian-main} shows that GCV recovers most of the oracle gain.
It also illustrates the central sign rule: in over-regularized regimes $\xi^\star<0$, so SD constrained to $\xi\in[0,1]$ cannot improve on the teacher. At $\lambda\approx31.6$, for example, the teacher velocity risk is about $1.67$, oracle SD gives $1.14$, and one-shot SD gives $1.22$.

\paragraph{Nonlinear Gaussian mixtures.}
We next use a four-component Gaussian mixture whose nonlinear population RF velocity is computable and exactly representable by our oracle feature map.
Figure~\ref{fig:mog-scatter} shows that the same correction mechanism carries over to nonlinear generation.
At $\lambda\approx7.5$, the teacher risk is $2.077$, pure distillation worsens it to $2.639$, oracle SD reduces it to $0.265$, and GCV-SD gives $0.273$ with $\widehat\xi\approx-6.46$.
Sliced $\Wtwo^2$ improves from about $1.02$ for the teacher to $0.093$ for GCV-SD, while mode mass and covariance errors also improve.

\paragraph{Finite-step generation.}
We evaluate Euler sampling across a range of numbers of function evaluations (NFEs).
In the same Gaussian mixture setting, oracle and GCV-SD improve generation metrics across NFEs. At NFE $=32$, sliced $\Wtwo^2$ is about $1.18$ for the teacher, $0.096$ for oracle SD, and $0.091$ for one-shot SD; see Section~\ref{app:finite-step-sampling} for the full curves.
These results are consistent with the generation bounds: improving the velocity estimation term can improve finite-step sampling.

\paragraph{Controlled tests on real images.}
Figure~\ref{fig:intro-selling} reports controlled neural RF tests on real images.
The teacher output is deliberately scaled down to emulate excessive shrinkage or miscalibration of the velocity scale.
Pure distillation compounds the shrinkage, while unconstrained SD selects negative mixing and recovers recognizable samples.
For Fashion-MNIST, validation selects $\widehat\xi=-2.83$ and reduces RF risk from $0.828$ to $0.210$. The best Feature-FD improves from $818.5$ to $8.8$, and conditioning accuracy at NFE $=64$ improves from $0.107$ to $0.895$. At NFE $=32$, the same experiment improves Inception-FID from $355.1$ to $14.8$; see Table~\ref{tab:fashion-fid} in the supplement.

\paragraph{Neural CIFAR-10 experiment.}
We further test whether negative mixing remains useful beyond linear probing with fixed features. Starting from a trained time-conditioned U-Net RF field $v_T$ on $32\times32$ CIFAR-10, we use $v=a v_T$ with $a=0.75$ as the teacher, thereby introducing a controlled miscalibration of the velocity scale. We then train a pure-distilled U-Net and tune $\xi$ by validation over $v_\xi=v+\xi(v_{\pd}-v)$. Validation selects $\widehat\xi=-1.225$, so restricting $\xi$ to $[0,1]$ returns the teacher. Table~\ref{tab:cifar-neural-rebuttal} and Figure~\ref{fig:cifar-neural-main} show that optimal SD improves RF risk from $0.2435$ to $0.1803$, FID from $284.3$ to $30.1$, and KID from $0.3215$ to $0.0186$ at NFE $=32$, whereas pure distillation worsens the teacher. Table~\ref{tab:cifar-neural-scale-sweep} in Appendix~\ref{app:cifar-neural} shows the same pattern across output scales: as $a$ increases toward one and the miscalibration becomes milder, the selected negative correction moves toward zero and the gain shrinks. The appendix also reports a fine-tuning baseline with matched compute.

\section{Discussion}
\label{sec:discussion}

In this paper, we identify a one-shot mechanism by which an RF model can improve from its own predictions.
For linear RF trained on fixed interpolants, the self-distilled student follows an affine path from the teacher to its pure-distilled refit, and the geometry of squared risk yields an optimal mixing coefficient in closed form.
The resulting student strictly improves every nonstationary teacher along the regularization path.
Unlike recursive Reflow, our procedure does not replace endpoint pairs or change interpolation covariates; it optimizes only the mixture of true and teacher velocity targets.

The main guarantee concerns integrated RF velocity risk for a fixed teacher regularization level.
It does not claim that the teacher is globally optimally tuned, nor that every downstream finite-step generation metric must improve under arbitrary misspecification.
Rather, it shows that a suboptimal regularization level defines a teacher-to-pure-distillation direction along which velocity risk can be reduced.
Through RF Wasserstein bounds, this risk reduction also tightens natural upper bounds on generation error.

Empirically, pure distillation can worsen over-regularized teachers because it repeats the same shrinkage.
Optimal SD often corrects this by extrapolating in the opposite direction, yielding a negative mixing coefficient.
This need for extrapolation is the main practical distinction from conventional convex distillation.
In practice, one can tune $\xi$ by GCV or validation; when the estimated gain is small, the selected coefficient remains close to zero.

The strict improvement theorem is exact only for linear RF with ridge regularization; the neural affine family in \eqref{eq:validation-mixture} is therefore an empirical extension, and a corresponding guarantee for end-to-end nonlinear training remains open.
Our controlled experiments on real images are diagnostic stress tests of teacher miscalibration: the CIFAR-10 scale sweep in Appendix~\ref{app:cifar-neural} traces the correction across miscalibration levels, while the Fashion-MNIST sensitivity analysis shows that gains diminish as the teacher improves.
Finally, extending the method to recursive endpoint replacement would bring it closer to Reflow practice, but requires new tools because interpolation covariates change across rounds.

\begin{ack}
We thank Hien Dang, Alessandro Rinaldo, and Sujay Sanghavi for helpful conversations.
Computing support is in part provided by the Texas Advanced Computing Center (TACC).
\end{ack}

\bibliographystyle{plainnat}
\bibliography{references-sd-generative-clean-with-urls}

\clearpage
\appendix
\setcounter{figure}{0}
\setcounter{table}{0}
\renewcommand{\thefigure}{S\arabic{figure}}
\renewcommand{\thetable}{S\arabic{table}}

\begin{center}
\Large
{\bf Supplement}
\end{center}

This supplement accompanies the paper ``Optimal Self-Distillation for Rectified Flow via Linear Probing''.
It contains proofs, further theoretical remarks, and additional experimental details and plots.
The supplement is organized as follows.

\addcontentsline{toc}{section}{Appendix} 
\startcontents
\printcontents{}{1}{\setcounter{tocdepth}{2}}

\clearpage
\section{Proofs for Section~\ref{sec:sd-velocity-learning}}
\label{app:proofs}

\subsection{Proof of Proposition~\ref{prop:affine}}
For fixed $\lambda$, the ridge solution is linear in the response matrix. Define
\[
    Q_\lambda=\left(\frac{\Phi^\top\Phi}{N}+\lambda I_m\right)^{-1}.
\]
Using the definition of the mixed target matrix,
\begin{align*}
    \widehat W_{\sd,\lambda,\xi}
    &=Q_\lambda\frac{\Phi^\top\Umat^{(\xi)}}{N}\\
    &=(1-\xi)Q_\lambda\frac{\Phi^\top\Umat}{N}
      +\xi Q_\lambda\frac{\Phi^\top\Uhat_\lambda}{N}\\
    &=(1-\xi)\widehat W_\lambda+\xi\widehat W_{\pd,\lambda}.
\end{align*}
The prediction identity follows from the linearity of $v_W(t,x)=W^\top\phi(t,x)$ in $W$.

\subsection{Proof of Theorem~\ref{thm:optimal-sd}}
Let
\[
    r_T=u-v_{\widehat W_\lambda}(T,X_T),\qquad
    r_P=u-v_{\widehat W_{\pd,\lambda}}(T,X_T).
\]
By Proposition~\ref{prop:affine}, the self-distilled residual is $(1-\xi)r_T+\xi r_P$.
Expanding the squared norm and taking conditional expectation gives
\begin{align*}
    R_{\sd}(\lambda,\xi)
    &=(1-\xi)^2R(\lambda)+\xi^2R_{\pd}(\lambda)+2\xi(1-\xi)C(\lambda)\\
    &=R(\lambda)-2\xi\{R(\lambda)-C(\lambda)\}+\xi^2D(\lambda).
\end{align*}
Since $D(\lambda)>0$, completing the square yields
\[
    R_{\sd}(\lambda,\xi)
    =D(\lambda)\left\{\xi-\frac{R(\lambda)-C(\lambda)}{D(\lambda)}\right\}^2
    +R(\lambda)-\frac{\{R(\lambda)-C(\lambda)\}^2}{D(\lambda)},
\]
which proves both identities in \eqref{eq:optimal-xi}.

It remains to prove the derivative identity.
Set $\widehat\Sigma=\Phi^\top\Phi/N$.
Since
\[
    \widehat W_\lambda=(\widehat\Sigma+\lambda I_m)^{-1}\frac{\Phi^\top\Umat}{N},
\]
its derivative is
\[
    \partial_\lambda\widehat W_\lambda=-(\widehat\Sigma+\lambda I_m)^{-1}\widehat W_\lambda.
\]
The pure-distilled coefficient is
\[
    \widehat W_{\pd,\lambda}=(\widehat\Sigma+\lambda I_m)^{-1}\widehat\Sigma\widehat W_\lambda.
\]
Using $(\widehat\Sigma+\lambda I_m)^{-1}\widehat\Sigma=I_m-\lambda(\widehat\Sigma+\lambda I_m)^{-1}$ gives
\[
    \widehat W_\lambda-\widehat W_{\pd,\lambda}=-\lambda\partial_\lambda\widehat W_\lambda.
\]
Because the velocity field is linear in its coefficient matrix, it follows that
\[
    v_{\pd,\lambda}(t,x)-v_\lambda(t,x)
    =\lambda\partial_\lambda v_\lambda(t,x).
\]
Consequently,
\begin{align*}
    R(\lambda)-C(\lambda)
    &=\E\left[
        \inner{u-v_\lambda(T,X_T)}
        {v_{\pd,\lambda}(T,X_T)-v_\lambda(T,X_T)}
        \mid\cD_N
    \right]\\
    &=\lambda\E\left[
        \inner{u-v_\lambda(T,X_T)}
        {\partial_\lambda v_\lambda(T,X_T)}
        \mid\cD_N
    \right].
\end{align*}
Differentiating the conditional risk gives
\[
    R'(\lambda)
    =-2\E\left[
        \inner{u-v_\lambda(T,X_T)}
        {\partial_\lambda v_\lambda(T,X_T)}
        \mid\cD_N
    \right],
\]
Combining the last two displays gives $R(\lambda)-C(\lambda)=-(\lambda/2)R'(\lambda)$.
Substituting this identity into \eqref{eq:optimal-xi} proves \eqref{eq:sign-rule}. Since $\lambda>0$ and $D(\lambda)>0$, the risk reduction is strict whenever $R'(\lambda)\neq0$, and the sign of $\xi^\star(\lambda)$ is the opposite of the sign of $R'(\lambda)$.

\section{Proofs and Gaussian calculations for Section~\ref{sec:generative-consequences}}
\label{app:generation-bound-proofs}

\subsection{Proof of Proposition~\ref{prop:continuous-generation-bound}}
The Wasserstein convergence theorem for rectified flow in \citet{bansal2026on} gives, under the stated Lipschitz and well-posedness assumptions,
\[
    \cW_2^2(\rho_v,\rho_1)
    \le \exp(1+2L)\int_0^1
    \E\norm{v^\star(t,X_t)-v(t,X_t)}^2\,\dd t.
\]
The integral is $\cE(v)$. The orthogonal decomposition \eqref{eq:irreducible} gives
\[
    \cR(v\mid\cD_N)=\cE(v)+\mathfrak S,
    \qquad
    \mathfrak S:=\E\norm{u-v^\star(T,X_T)}^2,
\]
where $\mathfrak S$ is independent of $v$. Since the optimized affine family contains both the teacher ($\xi=0$) and pure distillation ($\xi=1$),
\[
    \cE(v_{\sd}^\star)
    \le \min\{\cE(v_\lambda),\cE(v_{\pd,\lambda})\}.
\]
Moreover, Theorem~\ref{thm:optimal-sd} gives
\[
    \cE(v_\lambda)-\cE(v_{\sd}^\star)=R(\lambda)-R_{\sd}^\star(\lambda)
    =\frac{\lambda^2}{4}\frac{(R'(\lambda))^2}{D(\lambda)}.
\]
Multiplying this gain by the common factor $\exp(1+2L)$ proves Proposition~\ref{prop:continuous-generation-bound}.

\subsection{Proof of Theorem~\ref{thm:w2-ode-disc}}
\label{sec:proof-generation-error-discrete}
\subsubsection{Euler Wasserstein bound}
\label{sec:w2-convergence-discrete}
All expectations below are conditional on $\cD_N$, with the conditioning suppressed for readability. Couple the exact RF trajectory and the Euler sampler through the same initial point:
\begin{equation}
    \label{eq:og-rf}
    \frac{\dd Z_t}{\dd t}=v^\star(t,Z_t),\qquad Z_0\sim\rho_0.
\end{equation}
On the uniform grid $t_i=i/K$ with $h=K^{-1}$, the Euler iterates are
\[
    \widehat Y_{t_{i+1}}
    =\widehat Y_{t_i}+h v(t_i,\widehat Y_{t_i}),\qquad
    \widehat Y_0=Z_0,\qquad i=0,\ldots,K-1.
\]
For $t\in[t_i,t_{i+1}]$, define the continuous Euler interpolation
\begin{equation}
    \label{eq:interpolation-proc}
    \bar Y_t
    :=\widehat Y_{t_i}+(t-t_i)v(t_i,\widehat Y_{t_i}),
    \qquad i=0,\ldots,K-1.
\end{equation}
Thus $\bar Y_{t_i}=\widehat Y_{t_i}$, $\bar Y_{t_{i+1}}=\widehat Y_{t_{i+1}}$, and $\dd\bar Y_t/\dd t=v(t_i,\widehat Y_{t_i})$. Let
\[
    e_t:=\bar Y_t-Z_t,\qquad
    \delta_t:=v(t_i,\widehat Y_{t_i})-v^\star(t,Z_t).
\]
For any $\alpha>0$, Young's inequality gives
\begin{equation}
\label{eq:ddt-decomp}
    \frac{\dd}{\dd t}\norm{e_t}_2^2
    =2\inner{e_t}{\delta_t}
    \le \alpha\norm{e_t}_2^2+\frac{1}{\alpha}\norm{\delta_t}_2^2.
\end{equation}
Applying the integrating factor $\exp\{-\alpha(t-t_i)\}$ and integrating over $[t_i,t_{i+1}]$ yields
\begin{align*}
    \norm{\widehat Y_{t_{i+1}}-Z_{t_{i+1}}}_2^2
    &\le e^{\alpha h}\norm{\widehat Y_{t_i}-Z_{t_i}}_2^2\\
    &\quad+\frac{e^{\alpha h}}{\alpha}
    \int_{t_i}^{t_{i+1}}\norm{\delta_t}_2^2\,\dd t.
\end{align*}
Define
\[
    \Delta_i:=e^{-\alpha t_i}\E\norm{\widehat Y_{t_i}-Z_{t_i}}_2^2.
\]
Because $\widehat Y_0=Z_0$, we have $\Delta_0=0$. Taking expectations in the preceding inequality gives
\[
    \Delta_{i+1}
    \le \Delta_i+\frac{e^{-\alpha t_i}}{\alpha}
    \int_{t_i}^{t_{i+1}}\E\norm{\delta_t}_2^2\,\dd t.
\]
Decompose the integral using $\norm{a+b+c}^2\le3\{\norm{a}^2+\norm{b}^2+\norm{c}^2\}$:
\begin{align*}
    \int_{t_i}^{t_{i+1}}\E\norm{\delta_t}_2^2\,\dd t
    \le 3\{(\Rone)+(\Rtwo)+(\Rthree)\},
\end{align*}
where
\begin{align*}
    (\Rone)
    &:=\int_{t_i}^{t_{i+1}}
      \E\norm{v^\star(t,Z_t)-v^\star(t_i,Z_{t_i})}_2^2\,\dd t,\\
    (\Rtwo)
    &:=\int_{t_i}^{t_{i+1}}
      \E\norm{v^\star(t_i,Z_{t_i})-v(t_i,Z_{t_i})}_2^2\,\dd t,\\
    (\Rthree)
    &:=\int_{t_i}^{t_{i+1}}
      \E\norm{v(t_i,Z_{t_i})-v(t_i,\widehat Y_{t_i})}_2^2\,\dd t.
\end{align*}

We bound these three terms in turn. For $(\Rone)$, the fundamental theorem of calculus and Cauchy--Schwarz give, for $t\in[t_i,t_{i+1}]$,
\begin{equation}
\label{eq:straightness-bound}
\begin{aligned}
    \E\norm{v^\star(t,Z_t)-v^\star(t_i,Z_{t_i})}_2^2
    &\le (t-t_i)\int_{t_i}^{t}
    \E\norm{\frac{\dd}{\dd\tau}v^\star(\tau,Z_\tau)}_2^2\,\dd\tau\\
    &\le h^2\gamma_i,
\end{aligned}
\end{equation}
where
\[
    \gamma_i:=\frac1h\int_{t_i}^{t_{i+1}}
    \E\norm{\frac{\dd}{\dd\tau}v^\star(\tau,Z_\tau)}_2^2\,\dd\tau.
\]
Therefore,
\[
    (\Rone)\le h^3\gamma_i\le h^3\gamma_{2,K}(\cZ^\star).
\]
Since $Z_{t_i}$ has the same law as $X_{t_i}$, \eqref{eq:error-definitions} gives
\[
    (\Rtwo)=h\eps_i^2(v).
\]
Finally, the $L$-Lipschitz property of $v(t_i,\cdot)$ implies
\[
    (\Rthree)
    \le hL^2\E\norm{Z_{t_i}-\widehat Y_{t_i}}_2^2
    =hL^2e^{\alpha t_i}\Delta_i.
\]
Combining these bounds gives the recursion
\begin{equation}
\label{eq:moment-recursion}
    \Delta_{i+1}
    \le \left(1+\frac{3L^2h}{\alpha}\right)\Delta_i
    +\frac{3e^{-\alpha t_i}}{\alpha}
    \left\{h^3\gamma_{2,K}(\cZ^\star)+h\eps_i^2(v)\right\}.
\end{equation}
Let $q:=1+3L^2h/\alpha$. Dividing \eqref{eq:moment-recursion} by $q^{i+1}$, summing over $i$, and using $e^{-\alpha t_i}/q^{i+1}\le1$ gives
\[
    \frac{\Delta_K}{q^K}
    \le \frac{3}{\alpha}\left\{
        \frac{\gamma_{2,K}(\cZ^\star)}{K^2}
        +\bar\eps_K^2(v)
    \right\}.
\]
Since $q^K\le\exp(3L^2/\alpha)$ and
\[
    \cW_2^2(\widehat\rho_{v,K},\rho_1)
    \le \E\norm{\widehat Y_{t_K}-Z_{t_K}}_2^2
    =e^\alpha\Delta_K,
\]
we obtain, for every $\alpha>0$,
\begin{equation}
\label{eq:euler-alpha-bound}
    \cW_2^2(\widehat\rho_{v,K},\rho_1)
    \le \frac{3\exp(3L^2/\alpha+\alpha)}{\alpha}
    \left\{\frac{\gamma_{2,K}(\cZ^\star)}{K^2}+\bar\eps_K^2(v)\right\}.
\end{equation}
The prefactor is minimized at
\[
    \alpha^\star=\frac{1+\sqrt{1+12L^2}}{2}.
\]
Writing $s:=\sqrt{1+12L^2}$, substitution into \eqref{eq:euler-alpha-bound} gives the sharper constant
\[
    \widetilde C_L:=\frac{6e^s}{1+s}
    \le \frac{6e^{1+s}}{1+s}=C_L.
\]
Thus \eqref{eq:euler-w2-bound} follows with the stated constant $C_L$.

\subsubsection{Gridwise optimal mixing}
\label{sec:optimal-mixing-discrete-sampler}
For a fixed grid point $t_i$, the same quadratic expansion as in Theorem~\ref{thm:optimal-sd} gives
\[
    R_{\sd,t_i}(\lambda,\xi)
    =R_{t_i}(\lambda)-2\xi A_{t_i}(\lambda)+\xi^2D_{t_i}(\lambda).
\]
The time-specific irreducible term
\[
    \mathfrak S_{t_i}
    :=\E\norm{u-v^\star(t_i,X_{t_i})}_2^2
\]
is independent of $v$, so the same quadratic and gain hold for $\cE_{t_i}$. If $D_{t_i}(\lambda)>0$, minimizing over a time-specific coefficient gives
\[
    \xi_{t_i}^\star(\lambda)
    =\frac{A_{t_i}(\lambda)}{D_{t_i}(\lambda)},\qquad
    \cE_{t_i}(v_\lambda)
    -\cE_{t_i}(v_{\sd,\lambda,\xi_{t_i}^\star})
    =\frac{A_{t_i}(\lambda)^2}{D_{t_i}(\lambda)}.
\]
Whenever $R_{t_i}(\lambda)$ is differentiable, the fixed-time version of the derivative identity in Appendix~\ref{app:proofs} gives
\[
    A_{t_i}(\lambda)=-\frac{\lambda}{2}\partial_\lambda R_{t_i}(\lambda).
\]

For a shared scalar $\xi$ across the grid, the average error is
\[
    \bar\eps_K^2(v_{\sd,\lambda,\xi})
    =\frac1K\sum_{i=0}^{K-1}
    \left\{R_{t_i}(\lambda)-2\xi A_{t_i}(\lambda)
    +\xi^2D_{t_i}(\lambda)-\mathfrak S_{t_i}\right\}.
\]
If $\sum_iD_{t_i}(\lambda)>0$, completing the square gives
\[
    \xi_{\mathrm U,K}^\star(\lambda)
    =\frac{\sum_iA_{t_i}(\lambda)}{\sum_iD_{t_i}(\lambda)},
\]
and
\[
    \bar\eps_K^2(v_\lambda)-\bar\eps_K^2(v_{\sd,K}^\star)
    =\frac{\{\sum_iA_{t_i}(\lambda)\}^2}
    {K\sum_iD_{t_i}(\lambda)}.
\]
This proves \eqref{eq:grid-average-gain}. Because the shared affine family contains $\xi=0$ and $\xi=1$, its optimum is no worse than either the teacher or pure distillation. Combining this fact with the Wasserstein bound above and a common Lipschitz constant proves the remaining claims of Theorem~\ref{thm:w2-ode-disc}.

\subsection{Exact terminal errors in the Gaussian linear model}
\label{app:gaussian}

This subsection makes explicit, in a Gaussian linear model, how velocity error translates into terminal distribution error.
Suppose
\begin{equation}
    X_0\sim\N(0,I_d),\qquad X_1\sim\N(0,\Sigma),\qquad
    X_0\perp X_1,
\end{equation}
where $\Sigma\succ0$.
For $u=X_1-X_0$ and $X_t=(1-t)X_0+tX_1$,
\[
    C_t\coloneqq\operatorname{Cov}(X_t)=(1-t)^2I_d+t^2\Sigma,
    \qquad
    \operatorname{Cov}(u,X_t)=t\Sigma-(1-t)I_d.
\]
The Gaussian conditioning formula therefore gives
\begin{equation}
    \label{eq:gaussian-velocity}
    v^\star(t,x)=A_t^\star x,
    \qquad
    A_t^\star=\{t\Sigma-(1-t)I_d\}C_t^{-1}.
\end{equation}

Now consider a learned linear velocity $v(t,x)=A_tx$ whose matrices are jointly diagonalizable with $\Sigma$:
\[
    \Sigma=V\diag(s_1,\ldots,s_d)V^\top,
    \qquad
    A_t=V\diag(a_1(t),\ldots,a_d(t))V^\top.
\]
In this basis, write $a_j^\star(t)$ for the corresponding coefficient of $A_t^\star$, and define
\[
    \Delta_j(t)=a_j(t)-a_j^\star(t),
    \qquad
    M_j=\int_0^1\Delta_j(t)\,\dd t.
\]
Under the learned ODE, the variance $p_j(t)$ in coordinate $j$ satisfies
\[
    \frac{\dd p_j(t)}{\dd t}=2a_j(t)p_j(t),
    \qquad p_j(0)=1.
\]
The target flow has terminal variance $s_j$, so
\[
    p_j(1)
    =\exp\left\{2\int_0^1a_j(t)\,\dd t\right\}
    =s_je^{2M_j}.
\]
Thus the learned terminal covariance is
\begin{equation}
    P_1=V\diag\{s_1e^{2M_1},\ldots,s_de^{2M_d}\}V^\top.
\end{equation}
Because $P_1$ and $\Sigma$ share the same eigenvectors, the Gaussian Wasserstein and KL formulas reduce to
\begin{align}
    \cW_2^2(\N(0,P_1),\N(0,\Sigma))
    &=\sum_{j=1}^d s_j(e^{M_j}-1)^2,\label{eq:gauss-w2}\\
    \KL(\N(0,P_1)\,\|\,\N(0,\Sigma))
    &=\frac12\sum_{j=1}^d\{e^{2M_j}-1-2M_j\}.
    \label{eq:gauss-kl}
\end{align}

Finally, the reducible component of RF regression risk is the integrated squared velocity error along the true marginals:
\begin{equation}
    \int_0^1\E\!\left[\|v(t,X_t)-v^\star(t,X_t)\|^2\right]\dd t
    =\sum_{j=1}^d\mathcal{E}_j,
    \qquad
    \mathcal{E}_j\coloneqq\int_0^1c_j(t)\Delta_j(t)^2\,\dd t,
    \label{eq:gauss-risk}
\end{equation}
where $c_j(t)=(1-t)^2+t^2s_j$.
Weighted Cauchy--Schwarz yields
\begin{equation}
    M_j^2
    \leq\left(\int_0^1\frac{\dd t}{c_j(t)}\right)\mathcal{E}_j
    =\frac{\pi}{2\sqrt{s_j}}\,\mathcal{E}_j.
    \label{eq:gauss-cs}
\end{equation}
Equations~\eqref{eq:gauss-w2}--\eqref{eq:gauss-cs} therefore connect the exact terminal discrepancies to the same coordinatewise errors that comprise the RF excess risk.
These identities underlie the Gaussian experiments in Section~\ref{sec:experiments}.

\clearpage
\section{Additional experiments and details}
\label{app:additional-experiments}

This appendix supplements Section~\ref{sec:experiments} with detailed metrics and sensitivity checks for the image experiments, followed by additional studies of anisotropy, discretization, and regularization in Gaussian and Gaussian mixture models.

\subsection{Controlled experiments with handwritten digits and Fashion-MNIST}

\begin{figure*}[!ht]
    \centering
    \begin{subfigure}[t]{0.49\textwidth}
        \centering
        \includegraphics[width=\textwidth]{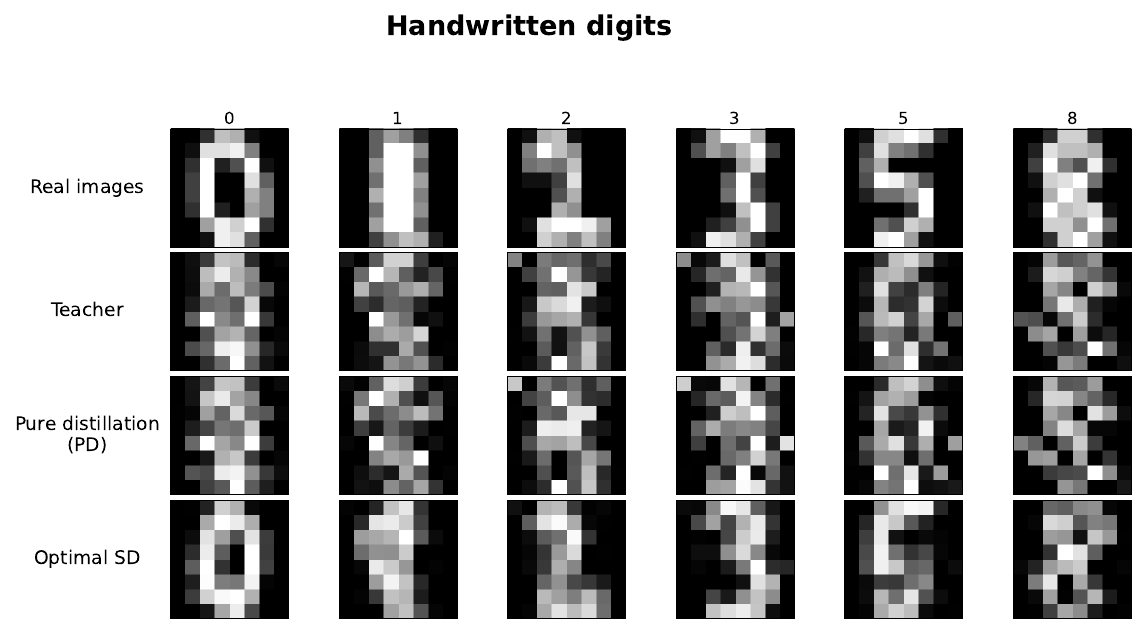}
        \caption{Handwritten digits.}
    \end{subfigure}
    \hfill
    \begin{subfigure}[t]{0.49\textwidth}
        \centering
        \includegraphics[width=\textwidth]{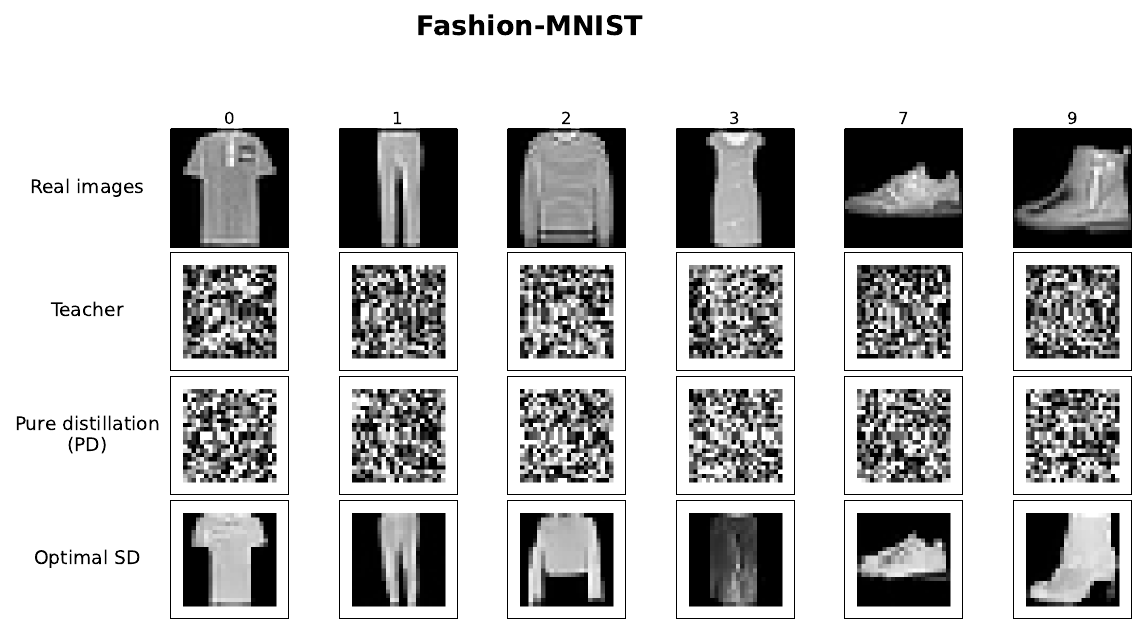}
        \caption{Fashion-MNIST.}
    \end{subfigure}
    \caption{\textbf{Results by dataset underlying Figure~\ref{fig:intro-selling}.}
    The panels show additional generated samples from the experiments with controlled shrinkage.}
    \label{fig:supp-split-selling}
\end{figure*}

\begin{figure*}[!ht]
    \centering
    \includegraphics[width=0.98\textwidth]{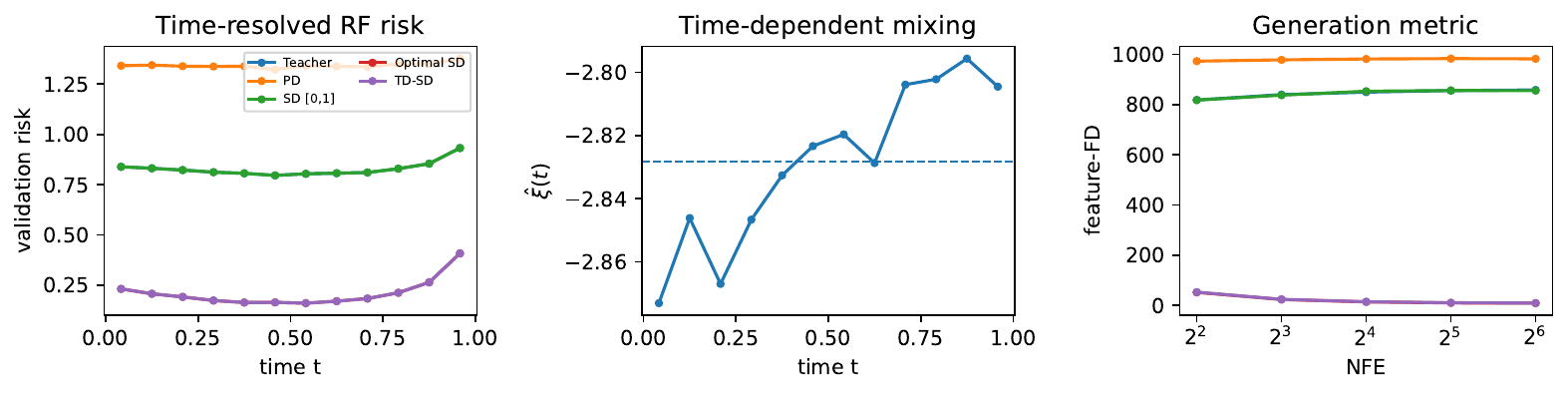}
    \caption{\textbf{Diagnostics for the controlled Fashion-MNIST stress test.}
    Left: time-resolved validation RF risk $R_t$.
    Middle: time-dependent mixing coefficient $\widehat\xi(t)$ tuned by validation, with the uniform coefficient shown as a dashed line.
    Right: Feature-FD versus number of Euler function evaluations.
    Clipped SD coincides with the teacher because the optimal correction is negative, while unconstrained and time-dependent SD reduce the velocity error terms that enter RF generation bounds.}
    \label{fig:supp-fashion-diagnostics}
\end{figure*}

\clearpage
\subsection{Metrics and sensitivity to teacher quality}
\label{app:metric-details}
For the controlled Fashion-MNIST experiment, Feature-FD is the Fr\'echet distance between generated and held-out real samples in the feature space defined by the penultimate layer of a Fashion-MNIST classifier trained on real training images.
Inception-FID is computed after converting grayscale images to RGB and resizing them to $299\times299$.
Conditioning accuracy is the fraction of generated samples classified as the requested class, and confidence is the average classifier probability assigned to that class.
Table~\ref{tab:fashion-fid} uses $10{,}000$ generated samples and $10{,}000$ held-out real samples at NFE $=32$.

The Fashion-MNIST result in Figure~\ref{fig:intro-selling} deliberately uses a teacher with strong output shrinkage to isolate the correction from negative mixing.
Table~\ref{tab:teacher-quality} compares this stress test with an unscaled, naturally trained teacher.
The latter has much lower initial risk and generation error; accordingly, validation selects a correction closer to zero and yields only a modest gain, consistent with its interpretation as a local correction.

\begin{table}[!ht]
\centering
\caption{\textbf{Generation metrics for the Fashion-MNIST stress test with controlled shrinkage.}
Metrics are computed from $10{,}000$ generated samples and $10{,}000$ held-out real samples at NFE $=32$, matching Figure~\ref{fig:intro-selling}.
Inception-FID is computed after converting grayscale images to RGB and resizing to $299\times299$; Feature-FD uses the penultimate layer of a Fashion-MNIST classifier.}
\label{tab:fashion-fid}
\begin{tabular}{ccccc}
\toprule
Method & Feature-FD $\downarrow$ & Inception-FID $\downarrow$ & Acc. $\uparrow$ & Conf. $\uparrow$ \\
\midrule
Teacher & 854.18 & 355.08 & 0.109 & 0.111 \\
Pure distillation & 983.10 & 388.11 & 0.102 & 0.103 \\
SD clipped $[0,1]$ & 857.26 & 355.10 & 0.113 & 0.114 \\
Optimal SD & 9.65 & 14.84 & 0.898 & 0.863 \\
\bottomrule
\end{tabular}
\end{table}

\begin{table}[!ht]
\centering
\caption{\textbf{Fashion-MNIST sensitivity to teacher quality.}
The controlled shrinkage teacher is deliberately degraded and yields a large negative correction.
The unscaled, naturally trained teacher requires a much smaller correction and yields a modest gain.
Feature-FD values are the best over the evaluated NFE grid.}
\label{tab:teacher-quality}
\begin{tabular}{cccccc}
\toprule
Teacher regime & Scale & $\widehat\xi$ & $R_T$ & $R_{\rm SD}$ & Feature-FD$_T \to$ Feature-FD$_{\rm SD}$ \\
\midrule
Controlled shrinkage & $0.35$ & $-2.83$ & $0.828$ & $0.210$ & $818.5 \to 8.8$ \\
Unscaled teacher & $1.00$ & $0.239$ & $0.205$ & $0.205$ & $9.88 \to 9.38$ \\
\bottomrule
\end{tabular}
\end{table}

\clearpage
\subsection{Neural CIFAR-10 RF experiment}
\label{app:cifar-neural}
Using a fully neural RF model on CIFAR-10 at $32\times32$ resolution, we test the correction mechanism beyond linear probing with fixed features.
We first train a time-conditioned U-Net velocity field with the standard RF objective, then use its output scaled by $a=0.75$ as a teacher with controlled miscalibration of the velocity scale.
We train a pure-distilled U-Net on the scaled teacher velocities and select $\xi$ by validation over $v_\xi=v+\xi(v_{\pd}-v)$.
Metrics are computed from $10{,}000$ generated samples using Euler sampling at NFE $=32$.
FID and KID use Inception features with the preprocessing implemented by torch-fidelity.
Table~\ref{tab:cifar-neural-full} also includes a fine-tuning baseline with matched compute, trained for the same number of adaptation steps as pure distillation.

\begin{figure*}[!ht]
    \centering
    \includegraphics[width=0.99\textwidth]{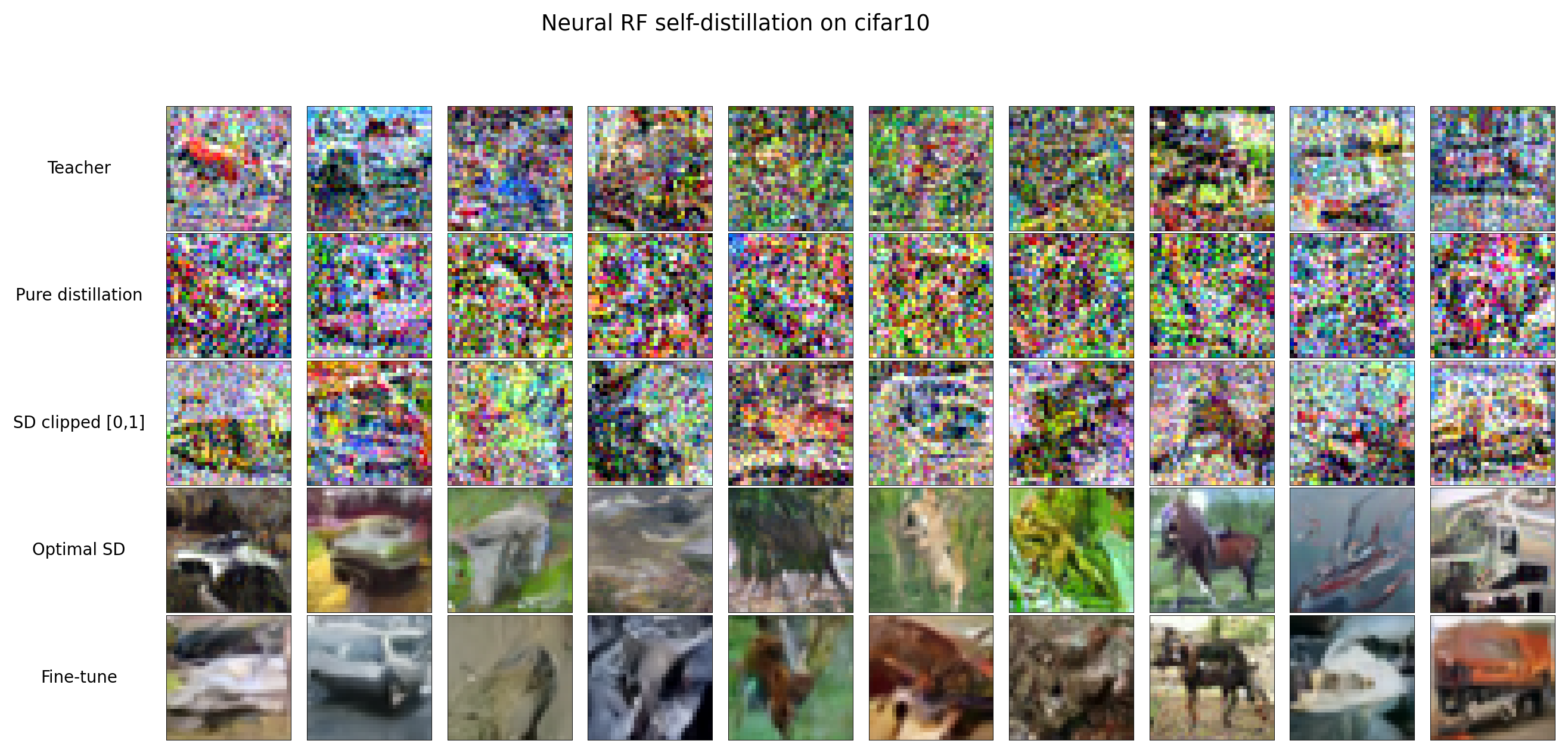}
    \caption{\textbf{Samples from the neural CIFAR-10 RF experiment.}
    We scale the output of a trained time-conditioned U-Net RF field by $a=0.75$ to introduce controlled miscalibration of the velocity scale.
    Pure distillation compounds the degradation, clipped SD coincides with the teacher because $\widehat\xi<0$, and unconstrained SD recovers substantially clearer samples.
    The final row shows the fine-tuning baseline with matched compute.}
    \label{fig:supp-cifar-neural-grid}
\end{figure*}

\begin{table}[!ht]
\centering
\small
\caption{\textbf{Neural CIFAR-10 RF experiment with a fine-tuning baseline.}
The baseline uses matched compute: it is initialized from the same scaled teacher and trained on true RF targets for the same number of adaptation steps as pure distillation.
Fine-tuning attains slightly lower RF risk, whereas optimal SD attains lower FID and KID.}
\label{tab:cifar-neural-full}
\begin{tabular}{lrrr}
\toprule
Method & RF risk $\downarrow$ & FID $\downarrow$ & KID $\downarrow$ \\
\midrule
Teacher & 0.2435 & 284.26 & 0.3215 \\
Pure distillation & 0.3885 & 324.77 & 0.3569 \\
SD clipped $[0,1]$ & 0.2435 & 283.56 & 0.3196 \\
Optimal SD & 0.1803 & 30.08 & 0.0186 \\
Compute-matched fine-tune & 0.1749 & 34.94 & 0.0268 \\
\bottomrule
\end{tabular}
\end{table}

We assess sensitivity to teacher output scaling in Table~\ref{tab:cifar-neural-scale-sweep}.
Across the evaluated scales, the correction selected by validation remains negative, pure distillation worsens the teacher, and optimal SD substantially improves FID.

\begin{table}[!ht]
\centering
\small
\caption{\textbf{Sensitivity of the neural CIFAR-10 experiment to teacher output scaling.}
Each row uses the same trained U-Net RF field scaled by $a$, $3{,}000$ generated samples, NFE $=32$, and the same adaptation budget.
Table~\ref{tab:cifar-neural-rebuttal} reports a separate evaluation with $10{,}000$ samples at $a=0.75$.}
\label{tab:cifar-neural-scale-sweep}
\begin{tabular}{lrrrr}
\toprule
Scale $a$ & $\widehat\xi$ & RF risk $T\to$ SD & FID $T\to$ SD & FID fine-tune \\
\midrule
$0.55$ & $-1.78$ & $0.396\to0.179$ & $324.4\to35.0$ & $47.0$ \\
$0.65$ & $-1.49$ & $0.310\to0.181$ & $321.1\to35.3$ & $46.9$ \\
$0.75$ & $-1.23$ & $0.245\to0.182$ & $291.9\to41.6$ & $46.9$ \\
$0.85$ & $-0.88$ & $0.201\to0.182$ & $212.4\to56.5$ & $47.0$ \\
\bottomrule
\end{tabular}

\end{table}

\begin{figure}[!ht]
    \centering
    \includegraphics[width=0.75\linewidth]{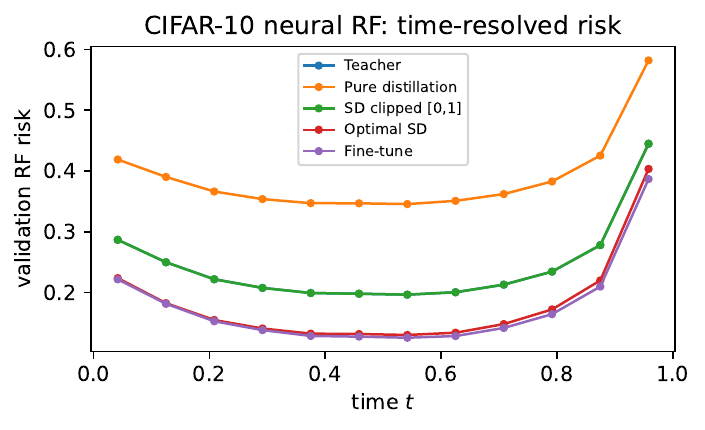}
    \caption{\textbf{Time-resolved validation RF risk for the neural CIFAR-10 RF experiment.}
    Optimal SD reduces RF risk relative to the scaled teacher and pure distillation across the time grid; clipped SD returns the teacher because validation selects a negative correction.}
    \label{fig:supp-cifar-neural-risk}
\end{figure}

\begin{figure}[!ht]
    \centering
    \includegraphics[width=0.75\linewidth]{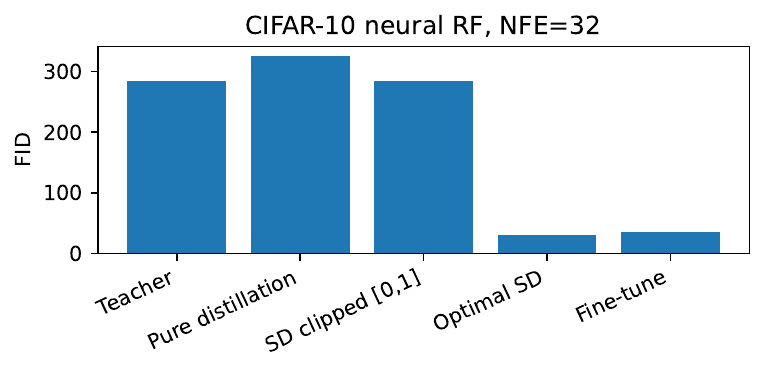}
    \caption{\textbf{FID comparison for the neural CIFAR-10 RF experiment.}
    At NFE $=32$, optimal SD substantially improves FID relative to the scaled teacher, pure distillation, clipped SD, and the fine-tuning baseline with matched compute.}
\label{fig:supp-cifar-neural-fid}
\end{figure}

\clearpage
\subsection{Anisotropic Gaussian experiment with temporal basis features}
\label{app:anisotropic}

\begin{figure}[!ht]
\centering
\includegraphics[width=0.99\linewidth]{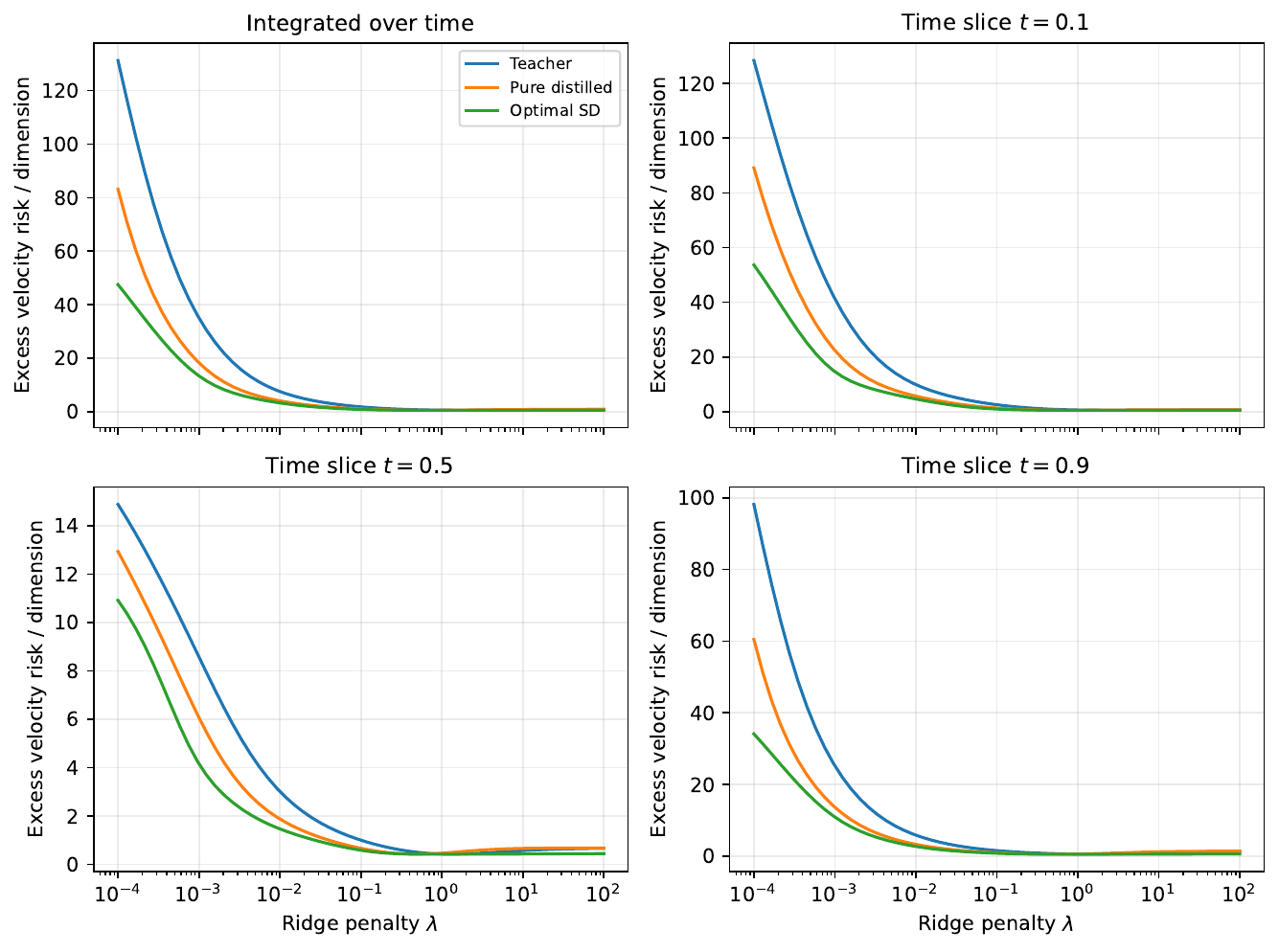}
\caption{
\textbf{Anisotropic Gaussian RF with temporal basis features.}
Optimal self-distillation improves both integrated velocity risk and the risk at individual times; over-regularized regimes require negative mixing.
}
\label{fig:anisotropic-app}
\end{figure}

We use an anisotropic Gaussian target with covariance eigenvalues logarithmically spaced between $0.15$ and $6$ and a model with a Legendre basis in time, $v_W(t,x)=\sum_{\ell=0}^5b_\ell(t)W_\ell x$.
Figure~\ref{fig:anisotropic-app} shows that the sign rule and strict improvement persist beyond the scalar Gaussian model with exact features.

\clearpage
\subsection{Nonlinear Gaussian mixture experiment}

\begin{figure}[!ht]
\centering
\includegraphics[width=0.99\linewidth]{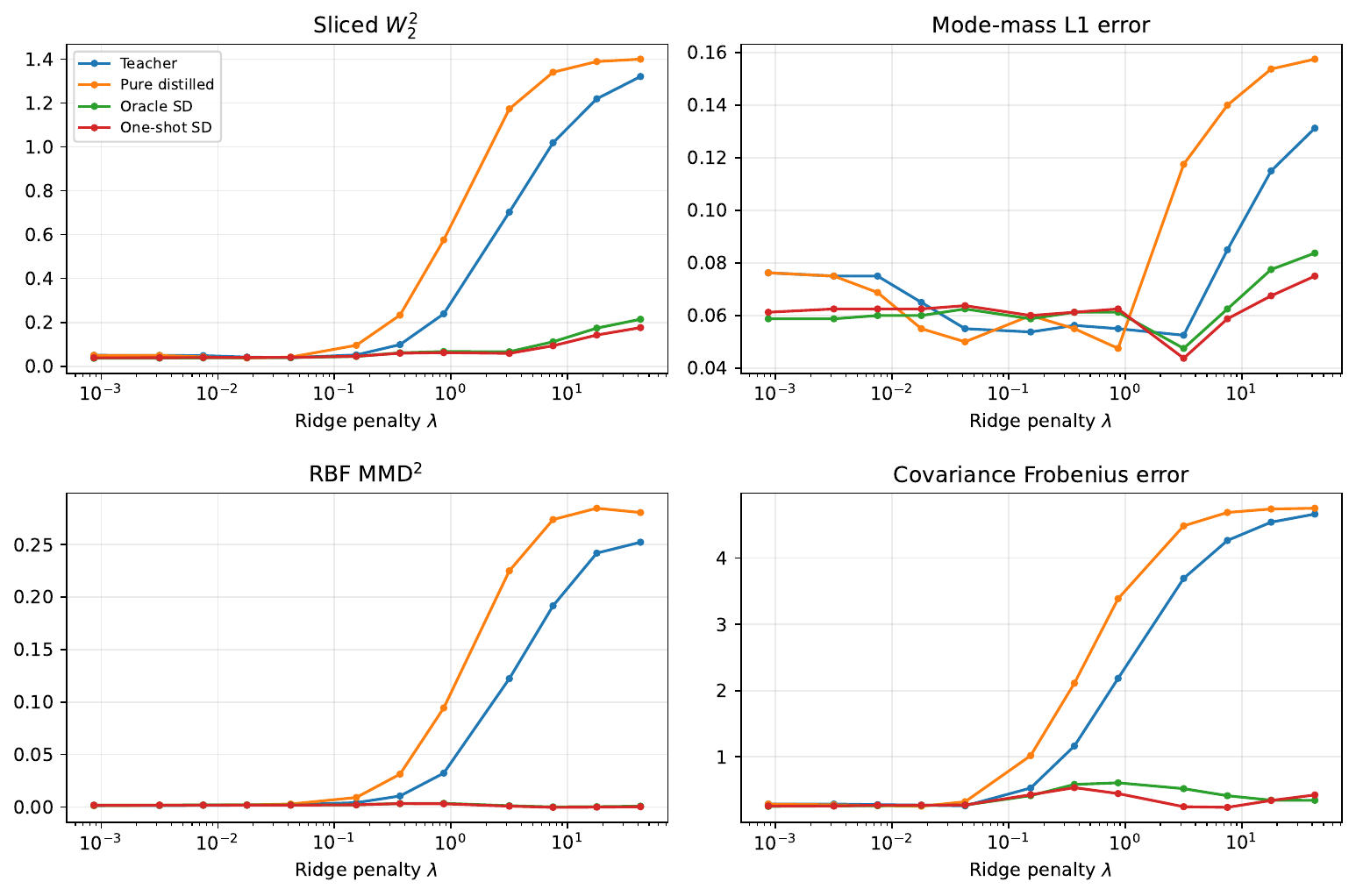}
\caption{
\textbf{Quantitative generation metrics for nonlinear Gaussian mixture RF.}
With oracle features, the MoG setting has a nonlinear population velocity but remains exactly representable.
Optimal self-distillation with negative mixing improves velocity risk and metrics of the generated distribution; the coefficient selected by GCV is close to the oracle coefficient.
}
\label{fig:mog-metrics}
\end{figure}

We use a four-component Gaussian mixture (MoG) in two dimensions.
The population velocity $\E[u\mid X_t=x]$ is nonlinear but computable from posterior responsibilities; we use an oracle feature map in which this velocity is exactly representable.
At $\lambda\approx7.5$, the teacher velocity risk is $2.08$, pure distillation worsens it to $2.64$, while oracle SD and one-shot SD reduce the risk to $0.265$ and $0.273$, respectively.
Figure~\ref{fig:mog-metrics} shows that the same correction improves sliced Wasserstein distance, maximum mean discrepancy (MMD), mode mass error, and covariance error.
Figure~\ref{fig:mog-scatter} visualizes the effect: the teacher produces a blurred mixture, while optimal SD with negative mixing recovers the four modes.

\clearpage
\subsection{Finite-step sampling in the Gaussian mixture model}
\label{app:finite-step-sampling}

\begin{figure}[!ht]
\centering
\includegraphics[width=0.99\linewidth]{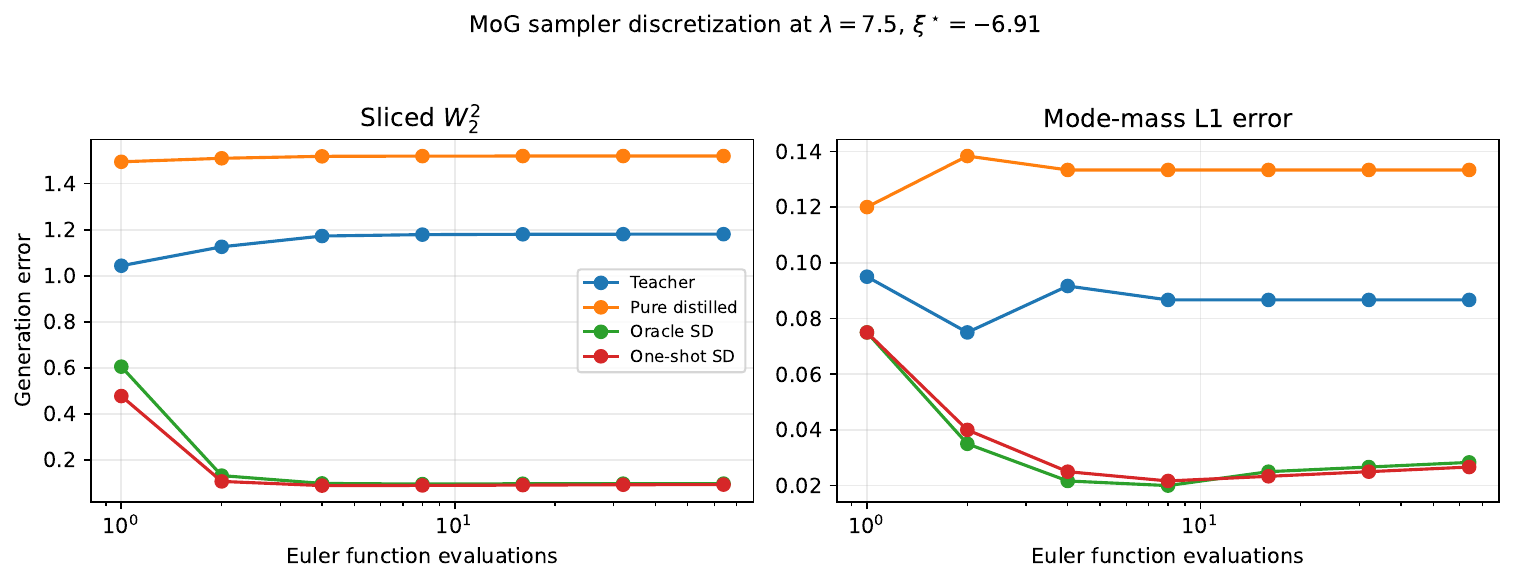}
\caption{
\textbf{Finite-step generation in nonlinear Gaussian mixture RF.}
Self-distillation improves generation metrics across Euler function evaluations, showing that the velocity correction persists under discretized RF sampling.
}
\label{fig:mog-nfe}
\end{figure}

Because RF generation requires numerical ODE integration, we evaluate Euler sampling over a range of numbers of function evaluations (NFEs).
This experiment examines whether the improvement in velocity risk translates into improved finite-step generation under the Euler discretization addressed by \eqref{eq:euler-w2-bound}.
In the MoG experiment with oracle features, SD improves generation metrics across NFEs.
At NFE $=32$, sliced $\cW_2^2$ is approximately $1.18$ for the teacher, $1.52$ for pure distillation, $0.096$ for oracle SD, and $0.091$ for one-shot SD; mode mass error drops from about $0.087$ for the teacher to $0.025$ for one-shot SD.
The corresponding curves appear in Figure~\ref{fig:mog-nfe}.

\subsection{Gaussian RF phase diagram across aspect ratios and ridge penalties}
\label{app:phase}

\begin{figure}[!ht]
\centering
\includegraphics[width=0.99\linewidth]{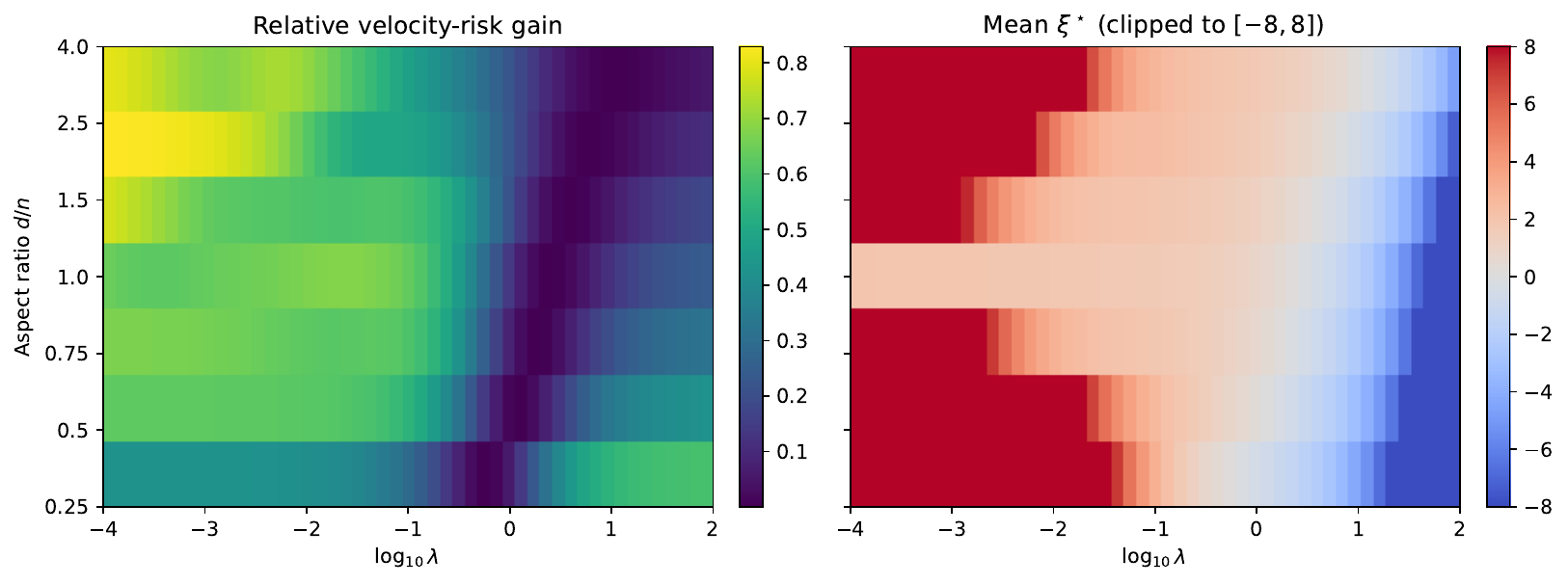}
\caption{
\textbf{Phase diagram of optimal self-distillation in Gaussian RF with exact features.}
Relative gain and sign of $\xi^\star$ vary systematically with aspect ratio and ridge penalty.
}
\label{fig:phase-app}
\end{figure}

Figure~\ref{fig:phase-app} varies the aspect ratio $d/n$ and regularization level in Gaussian RF with exact features.
The largest relative gains occur away from the optimal ridge region for the teacher, and the sign of $\xi^\star$ changes across the boundary between under- and over-regularization.

\clearpage
\section{Time-dependent optimal mixing}
\label{sec:time-varying-mixing}
\subsection{Pointwise and uniform coefficients}
\label{subsec:time-dependent}

Theorem~\ref{thm:optimal-sd} optimizes a single mixing coefficient shared across time.
Allowing a measurable function $t\mapsto\xi_t$ instead gives the time-dependent affine family
\begin{equation}
    v_{\sd,\lambda,\xi(\cdot)}(t,x)
    =v_\lambda(t,x)+\xi_t\{v_{\pd,\lambda}(t,x)-v_\lambda(t,x)\}.
    \label{eq:td-affine}
\end{equation}
Let $R_t(\lambda)$, $C_t(\lambda)$, and $D_t(\lambda)$ denote the analogues of \eqref{eq:C}--\eqref{eq:D} at a fixed time $t$, and set $A_t(\lambda)=R_t(\lambda)-C_t(\lambda)$.
Then the time-slice risk is
\begin{equation}
    R_{\sd,t}(\lambda,\xi_t)
    =R_t(\lambda)-2\xi_t A_t(\lambda)+\xi_t^2D_t(\lambda).
\end{equation}
Whenever $D_t(\lambda)>0$, the pointwise optimum is $\xi_t^\star(\lambda)=A_t(\lambda)/D_t(\lambda)$.
If $\int_0^1D_t(\lambda)\,\dd t>0$, the best coefficient shared across time is
\begin{equation}
    \xi_{\mathrm U}^\star(\lambda)=\frac{\int_0^1A_t(\lambda)\,\dd t}{\int_0^1D_t(\lambda)\,\dd t}.
    \label{eq:td-xi}
\end{equation}
Assume now that $D_t(\lambda)>0$ for almost every $t$, and define
$w_\lambda(t)=D_t(\lambda)/\int_0^1D_s(\lambda)\,\dd s$.
The additional gain in integrated risk from pointwise rather than uniform mixing is
\begin{equation}
\label{eq:td-extra-gain}
    \int_0^1\frac{A_t(\lambda)^2}{D_t(\lambda)}\,\dd t
    -\frac{\left(\int_0^1A_t(\lambda)\,\dd t\right)^2}{\int_0^1D_t(\lambda)\,\dd t}
    =\left(\int_0^1D_t(\lambda)\,\dd t\right)
    \mathrm{Var}_{w_\lambda}\{\xi_t^\star(\lambda)\}\ge0.
\end{equation}
Thus pointwise mixing performs at least as well as the best uniform mixture, with strict improvement precisely when $\xi_t^\star(\lambda)$ is not $w_\lambda$-almost surely constant.
The gridwise analogue in Appendix~\ref{sec:optimal-mixing-discrete-sampler} yields the reduction in velocity error used in the Euler generation bound \eqref{eq:euler-w2-bound}.

\clearpage
\subsection{Numerical evaluation}
\label{app:time-dependent-experiments}

We evaluate this time-dependent extension on four targets in two dimensions---a Gaussian mixture with four components (MoG-4), checkerboard, moons, and spiral---using RF with fixed random features and fixed interpolants.
The affine identity in Proposition~\ref{prop:affine} applies directly to a coefficient shared across all training samples.
With time-varying coefficients, however, refitting on targets mixed according to each sample's time generally differs from applying the affine construction \eqref{eq:td-affine} directly to the fields.
We therefore distinguish \emph{TD-affine}, which evaluates \eqref{eq:td-affine}, from \emph{TD-refit}, which retrains on time-dependent mixed targets.

\begin{figure}[!ht]
\centering
\includegraphics[width=0.86\linewidth]{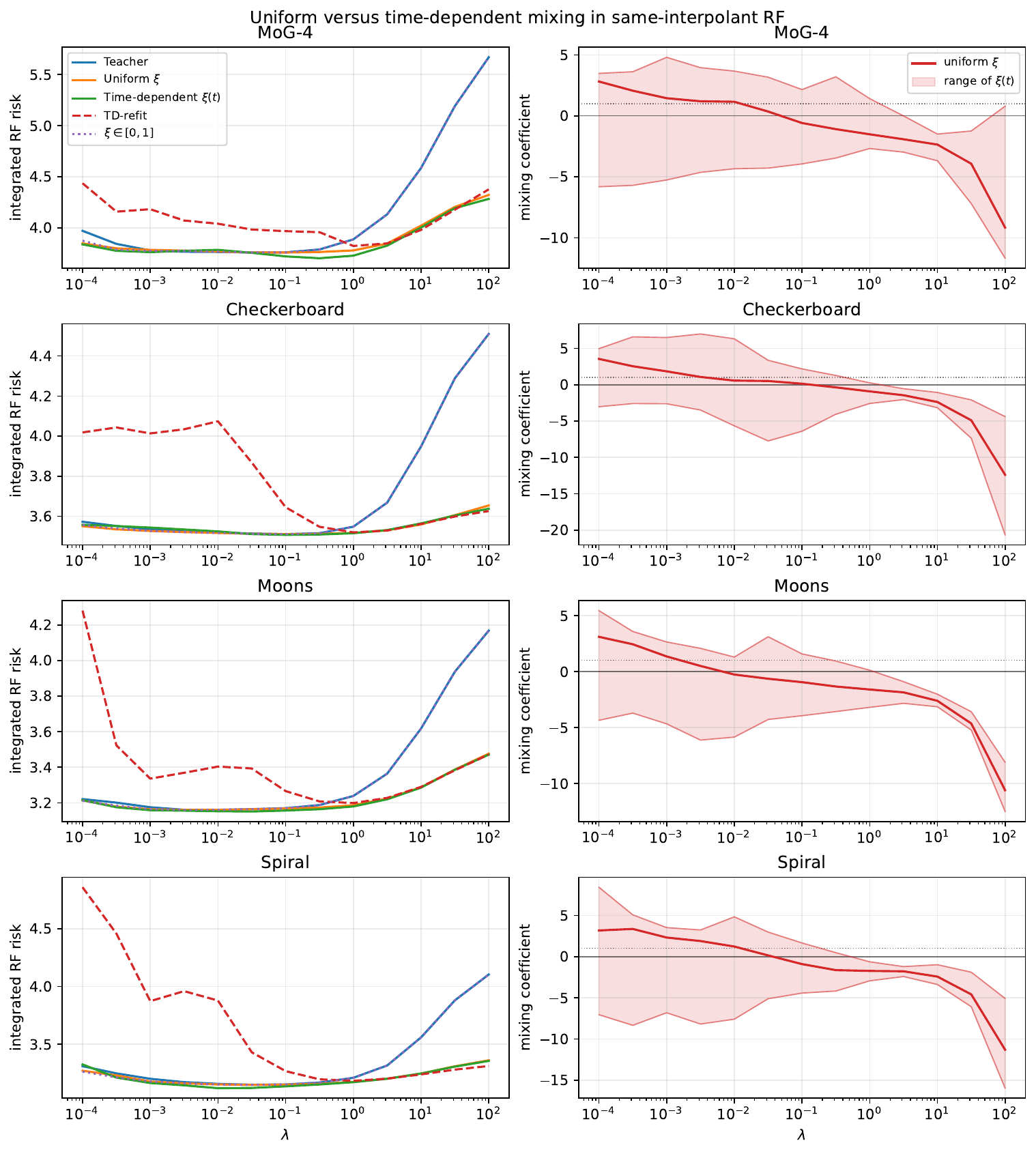}
\caption{
\textbf{Integrated RF risk under uniform and time-dependent mixing.}
Left: risk versus the teacher ridge penalty $\lambda$ for the teacher, the best uniform mixture, TD-affine, TD-refit, and uniform mixing constrained to $\xi\in[0,1]$.
Right: the optimal uniform coefficient (solid line) and the range of pointwise optimal coefficients $\xi_t^\star(\lambda)$ across time (shaded region).
TD-affine weakly improves on the best uniform mixture, as predicted by \eqref{eq:td-extra-gain}; TD-refit need not do so.
}
\label{fig:td-integrated-app}
\end{figure}

\begin{figure}[!ht]
\centering
\includegraphics[width=0.99\linewidth]{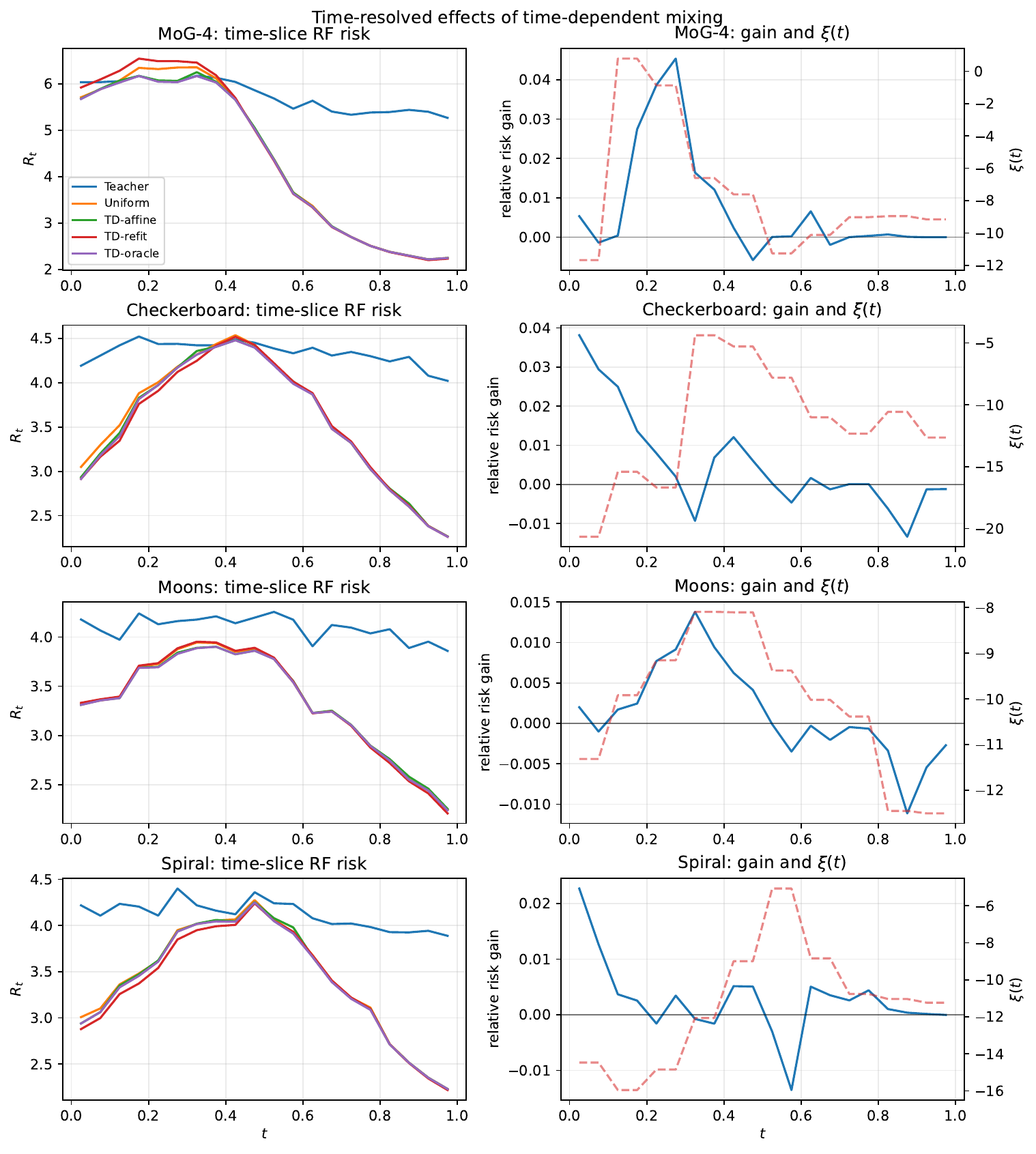}
\caption{
\textbf{Time-resolved RF risk and pointwise mixing coefficients.}
Left: risks at each time for the teacher, uniform mixing, TD-affine, TD-refit, and the pointwise oracle.
Right: the relative risk gain of TD-affine over uniform mixing (solid line; positive values favor TD-affine) and the selected $\xi_t$ (dashed line).
The gains concentrate at particular times, revealing where a shared coefficient under- or over-corrects.
}
\label{fig:td-timeslice-app}
\end{figure}

\begin{figure}[!ht]
\centering
\includegraphics[width=0.99\linewidth]{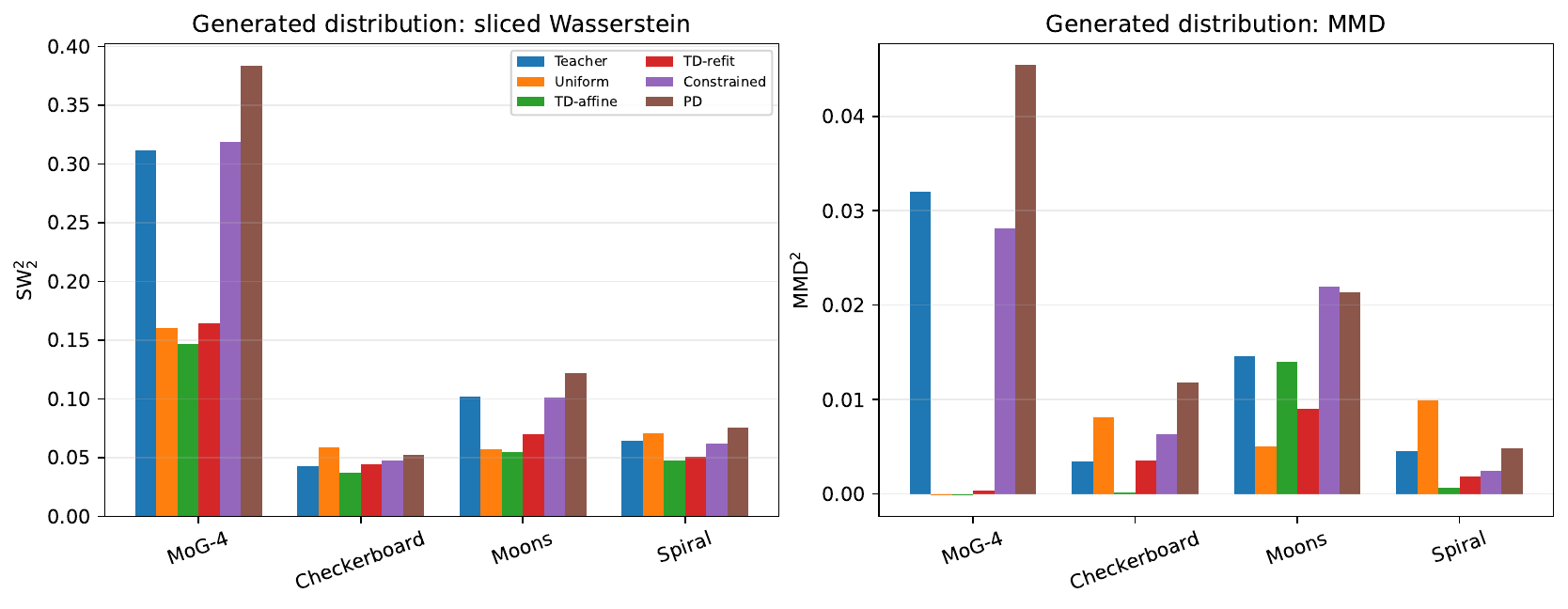}
\caption{
\textbf{Finite-step generation metrics under uniform and time-dependent mixing.}
Squared sliced Wasserstein distance (left) and squared MMD (right) for the four targets; lower is better.
The comparison includes the teacher, uniform mixing, TD-affine, TD-refit, mixing constrained to $[0,1]$, and pure distillation (PD).
Time-dependent mixing improves some target--metric pairs, but the guarantee on velocity risk does not imply uniform improvement in either generation metric.
}
\label{fig:td-generation-app}
\end{figure}

Figure~\ref{fig:td-integrated-app} compares their integrated RF risks with those of the teacher and the best uniform mixture across ridge penalties.
TD-affine provides modest, consistent improvements over the best uniform mixture, whereas TD-refit is not covered by \eqref{eq:td-extra-gain} and can perform worse.
Figure~\ref{fig:td-timeslice-app} localizes the gains across time, and Figure~\ref{fig:td-generation-app} reports the corresponding finite-step generation metrics.
The downstream improvements depend on the setting and metric, consistent with the scope of our guarantee on velocity risk.

\clearpage

\end{document}